# Online Speedup Learning for Optimal Planning

**Carmel Domshlak**                                      DCARMEL@IE.TECHNION.AC.IL
**Erez Karpas**                                           KARPASE@TECHNION.AC.IL
*Faculty of Industrial Engineering and Management*
*Technion - Israel Institute of Technology*
*Haifa, 32000, Israel*

**Shaul Markovitch**                                     SHAULM@CS.TECHNION.AC.IL
*Faculty of Computer Science*
*Technion - Israel Institute of Technology*
*Haifa, 32000, Israel*

## Abstract

Domain-independent planning is one of the foundational areas in the field of Artificial Intelligence. A description of a planning task consists of an initial world state, a goal, and a set of actions for modifying the world state. The objective is to find a sequence of actions, that is, a plan, that transforms the initial world state into a goal state. In optimal planning, we are interested in finding not just a plan, but one of the cheapest plans. A prominent approach to optimal planning these days is heuristic state-space search, guided by admissible heuristic functions. Numerous admissible heuristics have been developed, each with its own strengths and weaknesses, and it is well known that there is no single "best" heuristic for optimal planning in general. Thus, which heuristic to choose for a given planning task is a difficult question. This difficulty can be avoided by combining several heuristics, but that requires computing numerous heuristic estimates at each state, and the tradeoff between the time spent doing so and the time saved by the combined advantages of the different heuristics might be high. We present a novel method that reduces the cost of combining admissible heuristics for optimal planning, while maintaining its benefits. Using an idealized search space model, we formulate a decision rule for choosing the best heuristic to compute at each state. We then present an active online learning approach for learning a classifier with that decision rule as the target concept, and employ the learned classifier to decide which heuristic to compute at each state. We evaluate this technique empirically, and show that it substantially outperforms the standard method for combining several heuristics via their pointwise maximum.

## 1. Introduction

At the center of the problem of intelligent autonomous behavior is the task of *selecting the actions to take next*. Planning in AI is best conceived as the model-based approach to automated action selection (Geffner, 2010). The models represent the current situation, goals, and possible actions. Planning-specific languages are used to describe such models concisely. The main challenge in planning is computational, as most planning languages lead to intractable problems in the worst case. However, using rigorous search-guidance tools often allows for efficient solving of interesting problem instances.

In classical planning, which is concerned with the synthesis of plans constituting goal-achieving sequences of deterministic actions, significant algorithmic progress has been achieved in the last two decades. In turn, this progress in classical planning is translated to advances in more involved planning languages, allowing for uncertainty and feedback (Yoon, Fern, & Givan, 2007; Palacios

& Geffner, 2009; Keyder & Geffner, 2009; Brafman & Shani, 2012). In optimal planning, the objective is not just to find any plan, but to find one of the cheapest plans.

A prominent approach to domain-independent planning, and to optimal planning in particular, is state-space heuristic search. It is very natural to view a planning task as a search problem, and use a heuristic search algorithm to solve it. Recent advances in automatic construction of heuristics for domain-independent planning established many heuristics to choose from, each with its own strengths and weaknesses. However, this wealth of heuristics leads to a new question: given a specific planning task, which heuristic to choose?

In this paper, we propose *selective max* — an online learning approach that combines the strengths of several heuristic functions, leading to a speedup in optimal heuristic-search planning. At a high level, selective max can be seen as a hyper-heuristic (Burke, Kendall, Newall, Hart, Ross, & Schulenburg, 2003) — a heuristic for choosing among other heuristics. It is based on the seemingly trivial observation that, for each state, there is one heuristic which is the "best" for that state. In principle, it is possible to compute several heuristics for each state, and then choose one according to the values they provide. However, heuristic computation in domain-independent planning is typically expensive, and thus computing several heuristic estimates for each state takes a long time. Selective max works by predicting for each state which heuristic will yield the "best" heuristic estimate, and computes only that heuristic.

As it is not always clear how to decide what the "best" heuristic for each state is, we first analyze an idealized model of a search space and describe how to choose there the best heuristic for each state in order to minimize the overall search time. We then describe an online active learning procedure that uses a decision rule formulated for the idealized model. This procedure constitutes the essence of selective max.

Our experimental evaluation, which we conducted using three state-of-the-art heuristics for domain-independent planning, shows that selective max is very effective in combining several heuristics in optimal search. Furthermore, the results show that using selective max results in a speedup over the baseline heuristic combination method, and that selective max is robust to different parameter settings. These claims are further supported by selective max having been a runner-up ex-aequo in the last International Planning Competition, IPC-2011 (García-Olaya, Jiménez, & Linares López, 2011).

This paper expands on the conference version (Domshlak, Karpas, & Markovitch, 2010) in several ways. First, we improve and expand the presentation of the selective max decision rule. Second, we explain how to handle non-uniform action costs in a principled way. Third, the empirical evaluation is greatly extended, and now includes the results from IPC-2011, as well as controlled experiments with three different heuristics, and an exploration of how the parameters of selective max affect its performance.

## 2. Previous Work

Selective max is a *speedup learning* system. In general, speedup learning is concerned with improving the performance of a problem solving system with experience. The computational difficulty of domain-independent planning has led many researchers to use speedup learning techniques in order to improve the performance of planning systems; for a survey of many of these, see the work of Minton (1994), Zimmerman and Kambhampati (2003), and Fern, Khardon, and Tadepalli (2011).

Speedup learning systems can be divided along several dimensions (Zimmerman & Kambhampati, 2003; Fern, 2010). Arguably the most important dimension is the phase in which learning takes place. An offline, or inter-problem, speedup learner analyzes the problem solver's performance on different problem instances in an attempt to formulate some rule which would not only improve this performance but would also generalize well to future problem instances. Offline learning has been applied extensively to domain-independent planning, with varying degrees of success (Fern et al., 2011). However, one major drawback of offline learning is the need for training examples — in our case, planning tasks from the domains of interest.

Learning can also take place online, during problem solving. An *online, or intra-problem, speedup learner* is invoked by the problem solver on a concrete problem instance the solver is working on, and it attempts to learn online, with the objective of improving the solver's performance on that specific problem instance being solved. In general, online learners are not assumed to be pre-trained on some other, previously seen problem instances; all the information they can rely on has to be collected during the process of solving the concrete problem instance they were called for. Online learning has been shown to be extremely helpful in propositional satisfiability (SAT) and general constraint satisfaction (CSP) solving, where nogood learning and clause learning are now among the essential components of any state-of-the-art solver (Schiex & Verfaillie, 1993; Marques-Silva & Sakallah, 1996; Bayardo Jr. & Schrag, 1997). Thus, indirectly, SAT- and CSP-based domain-independent planners already benefit from these online learning techniques (Kautz & Selman, 1992; Rintanen, Heljanko, & Niemelä, 2006). However, to the best of our knowledge, our work is the first application of online learning to optimal heuristic-search planning.

## 3. Background

A domain-independent planning task (or *planning task*, for short) consists of a description of an initial state, a goal, and a set of available operators. Several formalisms for describing planning tasks are in use, including STRIPS (Fikes & Nilsson, 1971), ADL (Pednault, 1989), and SAS$^+$ (Bäckström & Klein, 1991; Bäckström & Nebel, 1995). We describe the SAS$^+$ formalism, the one used by the Fast Downward planner (Helmert, 2006), on top of which we have implemented and evaluated selective max. Nothing, however, precludes using selective max in the context of other formalisms.

A SAS$^+$ planning task is given by a 4-tuple $\Pi = \langle V, A, s_0, G \rangle$. $V = \{v_1, \ldots, v_n\}$ is a set of *state variables*, each associated with a finite domain $dom(v_i)$. A complete assignment $s$ to $V$ is called a *state*. $s_0$ is a specified state called the *initial state*, and the *goal* $G$ is a partial assignment to $V$. $A$ is a finite set of *actions*. Each action $a$ is given by a pair $\langle \mathsf{pre}(a), \mathsf{eff}(a) \rangle$ of partial assignments to $V$ called *preconditions* and *effects*, respectively. Each action $a$ also has an associated cost $\mathcal{C}(a) \in \mathbb{R}^{0+}$. An action $a$ is applicable in a state $s$ iff $s \models \mathsf{pre}(a)$. Applying $a$ changes the value of each state variable $v$ to $\mathsf{eff}(a)[v]$ if $\mathsf{eff}(a)[v]$ is specified. The resulting state is denoted by $s[\![a]\!]$. We denote the state obtained from sequential application of the (respectively applicable) actions $a_1, \ldots, a_k$ starting at state $s$ by $s[\![\langle a_1, \ldots, a_k \rangle]\!]$. Such an action sequence is a plan if $s_0[\![\langle a_1, \ldots, a_k \rangle]\!] \models G$. In optimal planning, we are interested in finding one of the cheapest plans, where the cost of a plan $\langle a_1, \ldots, a_k \rangle$ is the sum of its constituent action costs $\sum_{i=1}^{k} \mathcal{C}(a_i)$.

A SAS$^+$ planning task $\Pi = \langle V, A, s_0, G \rangle$ can be easily seen as a state-space search problem whose states are simply complete assignments to the variables $V$, with transitions uniquely determined by the actions $A$. The initial and goal states are also defined by the initial state and goal of $\Pi$. An optimal solution for a state-space search problem can be found by using the $A^*$ search algorithm with an admissible heuristic $h$. A heuristic evaluation function $h$ assigns an estimate of the distance to the closest goal state from each state it evaluates. The length of a cheapest path from state $s$ to the goal is denoted by $h^*(s)$, and $h$ is called *admissible* if it never overestimates the true goal distance — that is, if $h(s) \leq h^*(s)$ for any state $s$. $A^*$ works by expanding states in the order of increasing $f(s) := g(s) + h(s)$, where $g(s)$ is the cost of the cheapest path from the initial state to $s$ known so far.

## 4. Selective Max as a Decision Rule

Many admissible heuristics have been proposed for domain-independent planning; these vary from cheap to compute yet not very accurate, to more accurate yet expensive to compute. In general, the more accurate a heuristic is, the fewer states would be expanded by $A^*$ when using it. As the accuracy of heuristic functions varies for different planning tasks, and even for different states of the same task, we may be able to produce a more robust optimal planner by combining several admissible heuristics. Presumably, each heuristic is more accurate, that is, provides higher estimates, in different regions of the search space. The simplest and best-known way for doing that is using the point-wise maximum of the heuristics in use at each state. Given $n$ admissible heuristics, $h_1, \ldots, h_n$, a new heuristic, $\max_h$, is defined by $\max_h(s) := \max_{1 \leq i \leq n} h_i(s)$. It is easy to see that $\max_h(s) \geq h_i(s)$ for any state $s$ and for any heuristic $h_i$. Thus $A^*$ search using $\max_h$ is expected to expand fewer states than $A^*$ using any individual heuristic. However, if we denote the time needed to compute $h_i$ by $t_i$, the time needed to compute $\max_h$ is $\sum_{i=1}^n t_i$.

As mentioned previously, selective max is a form of hyper-heuristic (Burke et al., 2003) that chooses which heuristic to compute at each state. We can view selective max as a decision rule $dr$, which is given a set of heuristics $h_1, \ldots, h_n$ and a state $s$, and chooses which heuristic to compute for that state. One natural candidate for such a decision rule is the heuristic which yields the highest, that is, most accurate, estimate:

$$dr_{max}(\{h_1, \ldots, h_n\}, s) := h_{\operatorname{argmax}_{1 \leq i \leq n} h_i(s)}.$$

Using this decision rule yields a heuristic which is as accurate as $\max_h$, while still computing only one heuristic per state — in time $t_{\operatorname{argmax}_{1 \leq i \leq n} h_i(s)}$.

This analysis, however, does not take into account the different computation times of the different heuristics. For instance, let $h_1$ and $h_2$ be a pair of admissible heuristics such that $h_2 \geq h_1$. A priori, it seems that using $h_2$ should always be preferred to using $h_1$ because the former should cause $A^*$ to expand fewer states. However, suppose that on a given planning task, $A^*$ expands 1000 states when guided by $h_1$ and only 100 states when guided by $h_2$. If computing $h_1$ for each state takes 10 ms, and computing $h_2$ for each state takes 1000 ms, then switching from $h_1$ to $h_2$ *increases* the overall search time. Using $\max_h$ over $h_1$ and $h_2$ only makes things worse, because $h_2 \geq h_1$, and thus computing the maximum simply wastes the time spent on computing $h_1$. It is possible, however, that computing $h_2$ for a few carefully chosen states, and computing $h_1$ for all other states, would result in expanding 100 states, while reducing the overall search time when compared to running $A^*$ with only $h_2$.

As this example shows, even given knowledge of the heuristics' estimates in advance, it is not clear what heuristic should be computed at each state when our objective is to minimize the overall search time. Therefore, we begin by formulating a decision rule for choosing between one of two heuristics, with respect to an idealized state-space model. Selective max then operates as an online active learning procedure, attempting to predict the outcome of that decision rule and choose which heuristic to compute at each state.

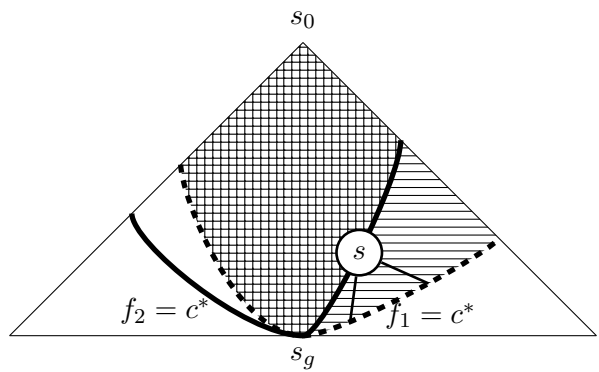

Figure 1: An illustration of the idealized search space model and the $f$-contours of two admissible heuristics

## 4.1 Decision Rule with Perfect Knowledge

We now formulate a decision rule for choosing which of two given admissible heuristics, $h_1$ and $h_2$, to compute for each state in an idealized search space model. In order to formulate such a decision rule, we make the following assumptions:

- The search space is a tree with a single goal, constant branching factor $b$, and uniform cost actions. Such an idealized search space model was used in the past to analyze the behavior of $A^*$ (Pearl, 1984).

- The time $t_i$ required for computing heuristic $h_i$ is independent of the state being evaluated; w.l.o.g. we assume $t_2 \geq t_1$.

- The heuristics are consistent. A heuristic $h$ is said to be consistent if it obeys the triangle inequality: For any two states $s, s'$, $h(s) \leq h(s') + k(s, s')$, where $k(s, s')$ is the optimal cost of reaching $s'$ from $s$.

- We have: (i) perfect knowledge about the structure of the search tree, and in particular the cost of the optimal solution $c^*$, (ii) perfect knowledge about the heuristic estimates for each state, and (iii) a perfect tie-breaking mechanism.

Obviously, none of the above assumptions holds in typical search problems, and later we examine their individual influence on our framework.

Adopting the standard notation, let $g(s)$ be the cost of the cheapest path from $s_0$ to $s$. Defining $\max_h(s) = \max(h_1(s), h_2(s))$, we then use the notation $f_1(s) = g(s) + h_1(s)$, $f_2(s) = g(s) + h_2(s)$, and $\max_f(s) = g(s) + \max_h(s)$. The $A^*$ algorithm with a consistent heuristic $h$ expands states in increasing order of $f = g + h$ (Pearl, 1984). In particular, every state $s$ with $f(s) < h^*(I) = c^*$ will surely be expanded by $A^*$, and every state with $f(s) > c^*$ will surely *not* be expanded by $A^*$. The states with $f(s) = c^*$ might or might not be expanded by $A^*$, depending on the tie-breaking rule being used. Under our perfect tie-breaking assumption, the only states with $f(s) = c^*$ that will be expanded are those that lie along some optimal plan.

Let us consider the states satisfying $f_1(s) = c^*$ (the dotted line in Fig. 1) and those satisfying $f_2(s) = c^*$ (the solid line in Fig. 1). The states above the $f_1 = c^*$ and $f_2 = c^*$ contours are those that are surely expanded by $A^*$ with $h_1$ and $h_2$, respectively. The states above both these contours (the grid-marked region in Fig. 1), that is, the states $SE = \{s \mid \max_f(s) < c^*\}$, are those that are surely expanded by $A^*$ using $\max_h$ (Pearl, 1984, Thm. 4, p. 79).

Under the objective of minimizing the search time, note that the optimal decision for any state $s \in SE$ is not to compute any heuristic at all, since all these states are surely expanded anyway. Assuming that we still must choose one of the heuristics, we would choose to compute the cheaper heuristic $h_1$. Another easy case is when $f_1(s) \geq c^*$. In these states, computing $h_1(s)$ suffices to ensure that $s$ is not surely expanded, and using a perfect tie-breaking rule, $s$ will not be expanded unless it must be. Because $h_1$ is also cheaper to compute than $h_2$, $h_1$ should be preferred, regardless of the heuristic estimate of $h_2$ for state $s$.

Let us now consider the optimal decision for all other states, that is, those with $f_1(s) < c^*$ and $f_2(s) \geq c^*$. In fact, it is enough to consider only the shallowest such states; in Figure 1, these are the states on the part of the $f_2 = c^*$ contour that separates between the grid-marked and line-marked areas. Since $f_1(s)$ and $f_2(s)$ are based on the same $g(s)$, we have $h_2(s) > h_1(s)$, that is, $h_2$ is more accurate in state $s$ than $h_1$. If we were interested solely in reducing state expansions, then $h_2$ would obviously be the right heuristic to compute at $s$. However, for our objective of reducing the actual search time, $h_2$ may actually be the wrong choice because it might be much more expensive to compute than $h_1$.

Let us consider the effects of each of our two alternatives. If we compute $h_2(s)$, then $s$ is not surely expanded, because $f_2(s) = c^*$, and thus whether or not $A^*$ expands $s$ depends on tie-breaking. As before, we are assuming perfect tie-breaking, and thus $s$ will not be expanded unless it must be. Computing $h_2$ would "cost" us $t_2$ time.

In contrast, if we compute $h_1(s)$, then $s$ is surely expanded because $f_1(s) < c^*$. Note that not computing $h_2$ for $s$ and then computing $h_2$ for one of the descendants $s'$ of $s$ is clearly a sub-optimal strategy as we do pay the cost of computing $h_2$, yet the pruning of $A^*$ is limited only to the search sub-tree rooted in $s'$. Therefore, our choices are really either computing $h_2$ for $s$, or computing $h_1$ for all the states in the sub-tree rooted in $s$ that lie on the $f_1 = c^*$ contour. Suppose we need to expand $l$ complete levels of the state space from $s$ to reach the $f_1 = c^*$ contour. Thus, we need to generate an order of $b^l$ states, and then invest $b^l t_1$ time in calculating $h_1$ for all these states that lie on the $f_1 = c^*$ contour.

Considering these two options, the optimal decision in state $s$ is thus to compute $h_2$ iff $t_2 < b^l t_1$, or to express it differently, if $l > \log_b(\frac{t_2}{t_1})$. As a special case, if both heuristics take the same time to compute, this decision rule reduces to $l > 0$, that is, the optimal choice is simply the more accurate heuristic for state $s$.

Putting all of the above cases together yields the decision rule $dr_{opt}$, as below, with $l_s$ being the depth to go from $s$ until $f_1(s) = c^*$:

$$dr_{opt}(\{h_1, h_2\}, s) := \begin{cases} h_1, & f_1(s) < c^*, f_2(s) < c^* \\ h_1, & f_1(s) \geq c^* \\ h_1, & f_1(s) < c^*, f_2(s) \geq c^*, l_s \leq \log_b(\frac{t_2}{t_1}) \\ h_2, & f_1(s) < c^*, f_2(s) \geq c^*, l_s > \log_b(\frac{t_2}{t_1}) \end{cases}.$$

## 4.2 Decision Rule without Perfect Knowledge

The idealized model above makes several assumptions, some of which appear to be very problematic to meet in practice. Here we examine these assumptions more closely, and when needed, suggest pragmatic compromises.

First, the model assumes that the search space forms a tree with a single goal state, that the heuristics in question are consistent, and that we have a perfect tie-breaking rule. Although the first assumption does not hold in most planning tasks, the second assumption is not satisfied by many state-of-the-art heuristics (Karpas & Domshlak, 2009; Helmert & Domshlak, 2009; Bonet & Helmert, 2010), and the third assumption is not realistic, they do not prevent us from using the decision rule suggested by the model.

The idealized model also assumes that both the branching factor and the heuristic computation times are constant across the search states. In our application of the decision rule to planning in practice, we deal with this assumption by adopting the average branching factor and heuristic computation times, estimated from a random sample of search states.

Finally, the decision rule $dr_{opt}$ above requires unrealistic knowledge of both heuristic estimates, as well as of the optimal plan cost $c^*$ and the depth $l_s$ to go from state until $f_1(s) = c^*$. As we obviously do not have this knowledge in practice, we must use some approximation of the decision rule.

The first approximation we make is to ignore the "trivial" cases that require knowledge of $c^*$; these are the cases where either $s$ is surely expanded, or $h_1$ is enough to prune $s$. Instead, we apply the reasoning for the "complicated" case for all states, resulting in the following decision rule:

$$dr_{app1}(\{h_1, h_2\}, s) := \begin{cases} h_1, & l_s \leq \log_b(\frac{t_2}{t_1}) \\ h_2, & l_s > \log_b(\frac{t_2}{t_1}) \end{cases}.$$

The next step is to somehow estimate the "depth to go" $l_s$ — the number of layers we need to expand in the tree until $f_1$ reaches $c^*$. In order to derive a useful decision rule, we assume that $l_s$ has a positive correlation with $\Delta_h(s) = h_2(s) - h_1(s)$; that is, if $h_1$ and $h_2$ are close, then $l_s$ is low, and if $h_1$ yields a much lower estimate than $h_2$, implying that $h_1$ is not very accurate for $s$, then the depth to go until $f_1(s) = c^*$ is large. Our approximation uses the simplest such correlation — a linear one — between $\Delta_h(s)$ and $l_s$, with a hyper-parameter $\alpha$ for controlling the slope.

Recall that in our idealized model, all actions were unit cost, and thus cost-to-go and depth-to-go are the same. However, some planning tasks, and notably, all planning tasks from the 2008 International Planning Competition, feature non-uniform action costs. Therefore, our decision rule converts heuristic estimates of cost-to-go into heuristic estimates of depth-to-go by dividing the cost-to-go estimate by the average action cost. We do this by modifying our estimate of the depth-to-go, $l_s$, with the average action cost, which we denote by $\hat{c}$. Plugging all of the above into our

decision rule yields:

$$dr_{app2}(\{h_1, h_2\}, s) := \begin{cases} h_1, & \Delta_h(s) \le \alpha \cdot \hat{c} \cdot \log_b(\frac{t_2}{t_1}) \\ h_2, & \Delta_h(s) > \alpha \cdot \hat{c} \cdot \log_b(\frac{t_2}{t_1}) \end{cases}.$$

Given $b, t_1, t_2$, and $\hat{c}$, the quantity $\alpha \cdot \hat{c} \cdot \log_b(t_2/t_1)$ becomes fixed, and in what follows we denote it simply by *threshold $\tau$*.

Note that linear correlation between $\Delta_h(s)$ and $l_s$ occurs in some simple cases. The first such case is when the $h_1$ value remains constant in the subtree rooted at $s$, that is, the additive error of $h_1$ increases by 1 for each level below $s$. In this case, $f_1$ increases by 1 for each expanded level of the sub-tree (because $h_1$ remains the same, and $g$ increases by 1), and it will take expanding exactly $\Delta_h(s) = h_2(s) - h_1(s)$ levels to reach the $f_1 = c^*$ contour. The second such case is when the absolute error of $h_1$ remains constant, that is, $h_1$ increases by 1 for each level expanded, and so $f_1$ increases by 2. In this case, we will need to expand $\Delta_h(s)/2$ levels. This can be generalized to the case where the estimate $h_1$ increases by any constant additive factor $c$, which results in $\Delta_h(s)/(c+1)$ levels being expanded.

Furthermore, there is some empirical evidence to support our conclusion about exponential growth of the search effort as a function of heuristic error, even when the assumptions made by the model do not hold. In particular, the experiments of Helmert and Röger (2008) on IPC benchmarks with heuristics with small constant additive errors show that the number of expanded nodes most typically grows exponentially as the (still very small and additive) error increases.

Finally, we remark that because our decision rule always chooses an admissible heuristic, the resulting heuristic estimate will always be admissible. Thus, even if the chosen heuristic is not the "correct" one according to $dr_{opt}$, this will not result in loss of optimality of the solution, but only in a possible increase in search time.

## 5. Online Learning of the Decision Rule

While decision rule $dr_{app2}$ still requires knowledge of $h_1$ and $h_2$, we can now use it as a binary label for each state. We can compute the value of the decision rule by "paying" the computation time of both heuristics, $t_1 + t_2$, and, more importantly, we can use a binary classifier to *predict* the value of this decision rule for some unknown state. Note that we use the classifier online, during the problem solving process, and the time spent on learning and classification is counted as time spent on problem solving. Furthermore, as in active learning, we can choose to "pay" for a label for some state, where the payment is also in computation time. Therefore we refer to our setting as *active online learning*.

In what follows, we provide a general overview of the selective max procedure, and describe several alternatives for each of its components. Our decision rule states that the more expensive heuristic $h_2$ should be computed at a search state $s$ when $h_2(s) - h_1(s) > \tau$. This decision rule serves as a binary target concept, which corresponds to the set of states where the more expensive heuristic $h_2$ is "significantly" more accurate than the cheaper heuristic $h_1$ — the states where, according to our model, the reduction in expanded states by computing $h_2$ outweighs the extra time needed to compute it. Selective max then uses a binary classifier to *predict* the value of the decision rule. There are several steps to building the classifier:

**evaluate**$(s)$
$\langle h, confidence \rangle :=$ CLASSIFY$(s, model)$
**if** $(confidence > \rho)$ **then**
   **return** $h(s)$
**else**
     $label := h_1$
     **if** $h_2(s) - h_1(s) > \alpha \cdot \hat{c} \cdot \log_b(t_2/t_1)$ **then** $label := h_2$
     update $model$ with $\langle s, label \rangle$
     **return** $\max(h_1(s), h_2(s))$

Figure 2: The *selective max* state evaluation procedure

1. *Training Example Collection*: We first need to collect training examples, which should be representative of the entire search space. Several state-space sampling methods are discussed in Section 5.1.

2. *Labeling Training Examples*: After the training examples are collected, they are first used to estimate the average branching factor $b$, average heuristic computation times $t_1$ and $t_2$, and the average action cost $\hat{c}$. Once $b, t_1, t_2,$ and $\hat{c}$ are estimated, we use them to estimate the threshold $\tau = \alpha \cdot \hat{c} \cdot \log_b(t_2/t_1)$ for the decision rule.

   We then generate a label for each training example by calculating $\Delta_h(s) = h_2(s) - h_1(s)$, and comparing it to the decision threshold: If $\Delta_h(s) > \tau$, we label $s$ with $h_2$, otherwise with $h_1$. If $t_1 > t_2$ we simply switch between the heuristics — our decision is always *whether or not to compute the more expensive heuristic*; the default is to compute the cheaper heuristic, unless the classifier says otherwise.

3. *Feature Extraction*: Having obtained a set of training examples, we must decide about the features to characterize each example. Since our target concept is based on heuristic values, the features should represent the information that heuristics are derived from — typically the problem description and the current state.

   While several feature-construction techniques for characterizing states of planning tasks have been proposed in previous literature (Yoon, Fern, & Givan, 2008; de la Rosa, Jiménez, & Borrajo, 2008), they were all designed for inter-problem learning, that is, for learning from different planning tasks which have already been solved offline. However, in our approach, we are only concerned with one problem, in an online setting, and thus these techniques are not applicable. In our implementation, we use the simplest features possible, taking each state variable as a feature. As our empirical evaluation demonstrates, even these elementary features suffice for selective max to perform well.

4. *Learning*: Once we have a set of labeled training examples, each represented by a vector of features, we can train a binary classifier. Several different choices of classifier are discussed in Section 5.2.

After completing the steps described above, we have a binary classifier that can be used to predict the value of our decision rule. However, as the classifier is not likely to have perfect accuracy, we further consult the confidence the classifier associates with its classification. The resulting state evaluation procedure of selective max is depicted in Figure 2. For every state $s$ evaluated by the search algorithm, we use our classifier to decide which heuristic to compute. If the classification confidence exceeds a confidence threshold $\rho$, a parameter of selective max, then only the indicated heuristic is computed for $s$. Otherwise, we conclude that there is not enough information to make a selective decision for $s$, and compute the regular maximum over $h_1(s)$ and $h_2(s)$. However, we use this opportunity to improve the quality of our prediction for states similar to $s$, and update our classifier by generating a label based on $h_2(s) - h_1(s)$ and learning from the newly labeled example. These decisions to dedicate computation time to obtain a label for a new example constitute the active part of our learning procedure. It is also possible to update the estimates for $b, t_1, t_2$, and $\hat{c}$, and change the threshold $\tau$ accordingly. However, this would result in the concept we are trying to learn constantly changing — a phenomenon known as *concept drift* — which usually affects learning adversely. Therefore, we do not update the threshold $\tau$.

### 5.1 State-Space Sampling

The initial state-space sample serves two purposes. First, it is used to estimate the branching factor $b$, the heuristic computation times $t_1$ and $t_2$, the average action cost $\hat{c}$, and then to compute the threshold $\tau = \alpha \cdot \hat{c} \cdot \log_b(t_2/t_1)$, which is used to specify our concept. After the concept is specified, the state-space sample also provides us with a set of examples on which the classifier is initially trained. Therefore, it is important to have an initial state-space sample that is representative of the states which will be evaluated during search. The number of states in the initial sample is controlled by a parameter $N$.

One option is to use the first $N$ states of the search. However, this method is biased towards states closer to the initial state, and therefore is not likely to represent the search space well. Thus, we discuss three more sophisticated state-space sampling procedures, all of which are based on performing random walks, or "probes," from the initial state. While the details of these sampling procedures vary, each such "probe" terminates at some pre-set depth limit.

The first sampling procedure, which we refer to as "biased probes," uses an inverse heuristic selection bias for choosing the next state to go to in the probe. Specifically, the probability of choosing state $s$ as the successor from which the random walk will continue is proportional to $1/\max_h(s)$. This biases the sample towards states with lower heuristic estimates, which are more likely to be expanded during the search.

The second sampling procedure is similar to the first one, except that it chooses the successor uniformly, and thus we refer to it as "unbiased probes." Both these sampling procedures add all of the generated states (that is, the states along the probe as well as their "siblings") to the state-space sample, and they both terminate after collecting $N$ training examples. The depth limit for all random walks is the same in both sampling schemes, and is set to some estimate of the goal depth; we discuss this goal depth estimate later.

The third state-space sampling procedure, referred to here as *PDB* sampling, has been proposed by Haslum, Botea, Helmert, Bonet, and Koenig (2007). This procedure also uses unbiased probes, but only adds the last state reached in each probe to the state-space sample. The depth of each probe is determined individually, by drawing a random depth from a binomial distribution around the estimated goal depth.

Note that all three sampling procedures rely on some estimate of the minimum goal depth. When all actions are unit cost, the minimum goal depth is the same as $h^*(s_0)$, and thus we can use a heuristic to estimate it. In our evaluation, we used twice the heuristic estimate of the initial state, $2 \cdot \max_h(s_0)$, as the goal depth estimate. However, with non-uniform action costs, goal depth and cost are no longer measured in the same units. While it seems we could divide the above heuristic-based estimate by the average action cost $\hat{c}$, recall that we use the state-space sample in order to obtain an estimate for estimate $\hat{c}$, thus creating a circular dependency. Although it is possible to estimate $\hat{c}$ by taking the average cost of all actions in the problem description, there is no reason to assume that all actions are equally likely to be used. Another option is to modify the above state-space sampling procedures, and place a cost limit, rather than a depth limit, on each probe. However, this would pose a problem in the presence of 0-cost actions. In such a case, when a probe reaches its cost limit yet has a possible 0-cost action to apply, it is not clear whether the probe should terminate. Therefore, we keep using depth-limited probes and attempt to estimate the depth of the cheapest goal. We compute a heuristic estimate for the initial state, and then use the *number* of actions which the heuristic estimate is based on as our goal depth estimate. While this is not possible with every heuristic, we use in our empirical evaluation the monotonically-relaxed plan heuristic. This heuristic, also known as the FF heuristic (Hoffmann & Nebel, 2001), does provide such information: we first use this heuristic to find a relaxed plan from the initial state, and then use the number of actions in the relaxed plan as our goal depth estimate.

## 5.2 Classifier

The last decision to be made is the choice of classifier. Although many classifiers can be used here, several requirements must be met due to our particular setup. First, both training and classification must be very fast, as both are performed during time-constrained problem solving. Second, the classifier must be incremental to support active learning. This is achieved by allowing online updates of the learned model. Finally, the classifier should provide us with a meaningful measure of confidence for its predictions.

While several classifiers meet these requirements, we found the Naive Bayes classifier to provide a good balance between speed and accuracy. One note on the Naive Bayes classifier is that it assumes a very strong conditional independence between the features. Although this is not a fully realistic assumption for planning tasks, using a SAS$^+$ task formulation in contrast to the classical STRIPS formulations helps a lot: instead of many highly dependent binary variables, we have a much smaller set of less dependent ones.

Although, as the empirical evaluation will demonstrate, Naive Bayes appears to be the most suitable classifier to use with selective max, other classifiers can also be used. The most obvious choice for a replacement classifier would be a different Bayesian classifier. One such classifier is AODE (Webb, Boughton, & Wang, 2005), an extension of Naive Bayes, which somewhat relaxes the assumption of independence between the features, and is typically more accurate than Naive Bayes. However, this added accuracy comes at the cost of increased training and classification time.

Decision trees are another popular type of classifier that allows for even faster classification. While most decision tree induction algorithms are not incremental, the Incremental Tree Inducer (ITI) algorithm (Utgoff, Berkman, & Clouse, 1997) supports incremental updating of decision trees by tree restructuring, and also has a freely available implementation in C. In our evaluation, we used ITI in incremental mode, and incorporated every example into the tree immediately, because the tree is likely to be used for many classifications between pairs of consecutive updates with training examples from active learning. The classification confidence with the ITI classifier is obtained by the frequency of examples at the leaf node from which the classification came.

A different family of possible classifiers is $k$-Nearest Neighbors (kNN) (Cover & Hart, 1967). In order to use kNN, we need a distance metric between examples, which, with our features, are simply states. As with our choice of features, we opt for simplicity and use Euclidean distance as our metric. kNN enjoys very fast learning time but suffers from slow classification time. The classification confidence is obtained by a simple (unweighted) vote between the $k$ nearest neighbors.

Another question related to the choice of classifier is feature selection. In some planning tasks, the number of variables, and accordingly, features, can be over 2000 (for example, task 35 of the AIRPORT domain has 2558 variables). While the performance of Naive Bayes and kNN can likely be improved using feature selection, doing so poses a problem when the initial sample is considered. Since feature selection will have to be done right after the initial sample is obtained, it will have to be based only on the initial sample. This could cause a problem since some features might appear to be irrelevant according to the initial sample, yet turn out to be very relevant when active learning is used after some low-confidence states are encountered. Therefore, we do not use feature selection in our empirical evaluation of selective max.

### 5.3 Extension to Multiple Heuristics

To this point, we have discussed how to choose which heuristic to compute for each state when there are only two heuristics to choose from. When given more than two heuristics, the decision rule presented in Section 4 is inapplicable, and extending it to handle more than two heuristics is not straightforward. However, extending selective max to use more than two heuristics is straightforward — simply compare heuristics in a pair-wise manner, and use a voting rule to choose which heuristic to compute.

While there are many possible such voting rules, we go with the simplest one, which compares every pair of heuristics, and chooses the winner by a vote, weighted by the confidence for each pairwise decision. The overall winner is simply the heuristic which has the highest total confidence from all pairwise comparisons, with ties broken in favor of the cheaper-to-compute heuristic. Although this requires a quadratic number of classifiers, training and classification time (at least with Naive Bayes) appear to be much lower than the overall time spent on heuristic computations, and thus the overhead induced by learning and classification is likely to remain relatively low for reasonable heuristic ensembles.

## 6. Experimental Evaluation

To evaluate selective max empirically, we implemented it on top of the open-source Fast Downward planner (Helmert, 2006). Our empirical evaluation is divided into three parts. First, we examine the performance of selective max using the last International Planning Competition, IPC-2011, as our benchmark. Selective max was the runner-up ex-aequo at IPC-2011, tying for 2nd place with a version of Fast Downward using an abstraction "merge-and-shrink" heuristic (Nissim, Hoffmann, & Helmert, 2011), and losing to a sequential portfolio combining the heuristics used in both runners-up (Helmert, Röger, & Karpas, 2011). Second, we present a series of controlled parametric experiments, where we examine the behavior of selective max under different settings. Finally, we compare selective max to a simulated sequential portfolio, using the same heuristics as selective max.

| Parameter | Default value | Meaning |
|-----------|---------------|---------|
| $\alpha$ | 1 | heuristic difference bias |
| $\rho$ | 0.6 | confidence threshold |
| $N$ | 1000 | initial sample size |
| Sampling method | Biased probes | state-space sampling method |
| Classifier | Naive Bayes | classifier type |

Table 1: Parameters for the *selmax* entry in IPC-2011.

## 6.1 Performance Evaluation: Results from IPC-2011

The IPC-2011 experiments (García-Olaya et al., 2011) were run by the IPC organizers, on their own machines, with a time limit of 30 minutes and a memory limit of 6 GB per planning task. The competition included some new domains, which none of the participants had seen before, thus precluding the participants from using offline learning approaches.

Although many planners participated in the sequential optimal track of IPC-2011, we report here only the results relevant to selective max. The selective max entry in IPC-2011 was called *selmax*, and consisted of selective max over the uniform action cost partitioning version of $h_{LA}$ (Karpas & Domshlak, 2009) and $h_{\text{LM-CUT}}$ (Helmert & Domshlak, 2009) heuristics. The parameters used for selective max in IPC-2011 are reported in Table 1. Additionally, each of the heuristics *selmax* used was entered individually as *BJOLP* ($h_{LA}$) and *lmcut* ($h_{\text{LM-CUT}}$), and we report results for all three planners. While a comparison of selective max with the regular maximum of $h_{LA}$ and $h_{\text{LM-CUT}}$ would be interesting, there was no such entry at IPC-2011, and thus we can not report on it. In our controlled experiments, we do compare selective max to the regular maximum, as well as to other baseline combination methods.

Figure 3 shows the anytime profile of these three planners on IPC-2011 tasks, plotting the number of tasks solved under different timeouts, up to the time limit of 30 minutes. Additionally, Table 2 shows the number of tasks solved in each domain of IPC-2011, after 30 minutes, and includes the number of problems solved by the winner, Fast Downward Stone Soup 1 (FDSS-1), for reference.

As these results show, selective max solves more problems than each of the individual heuristics it uses. Furthermore, the anytime profile of selective max dominates each of these heuristics, in the range between 214 seconds until the full 30 minute timeout. The behavior of the anytime plot with shorter timeouts is due to the overhead of selective max, which consists of obtaining the initial state-space sample, as well as learning and classification. However, it appears that selective max quickly compensates for its relatively slow start.

## 6.2 Controlled Experiments

In our series of controlled experiments, we attempted to evaluate the impact of different parameters on selective max. We controlled the following independent variables:

- *Heuristics:* We used three state-of-the-art admissible heuristics: $h_{LA}$ (Karpas & Domshlak, 2009), $h_{\text{LM-CUT}}$ (Helmert & Domshlak, 2009), and $h_{\text{LM-CUT}^+}$ (Bonet & Helmert, 2010). None of these base heuristics yields better search performance than the others across all planning domains. Of these heuristics, $h_{LA}$ is typically the fastest to compute and the least accurate, $h_{\text{LM-CUT}}$ is more expensive to compute and more accurate, and $h_{\text{LM-CUT}^+}$ is the most expensive to compute and the most accurate.[1] From the data we have gathered in these experiments, $h_{\text{LM-CUT}}$ takes on average 4.5 more time per state than $h_{LA}$, and $h_{\text{LM-CUT}^+}$ takes 53 more time per state than $h_{LA}$. We evaluate selective max with all possible subsets of two or more of these three heuristics.

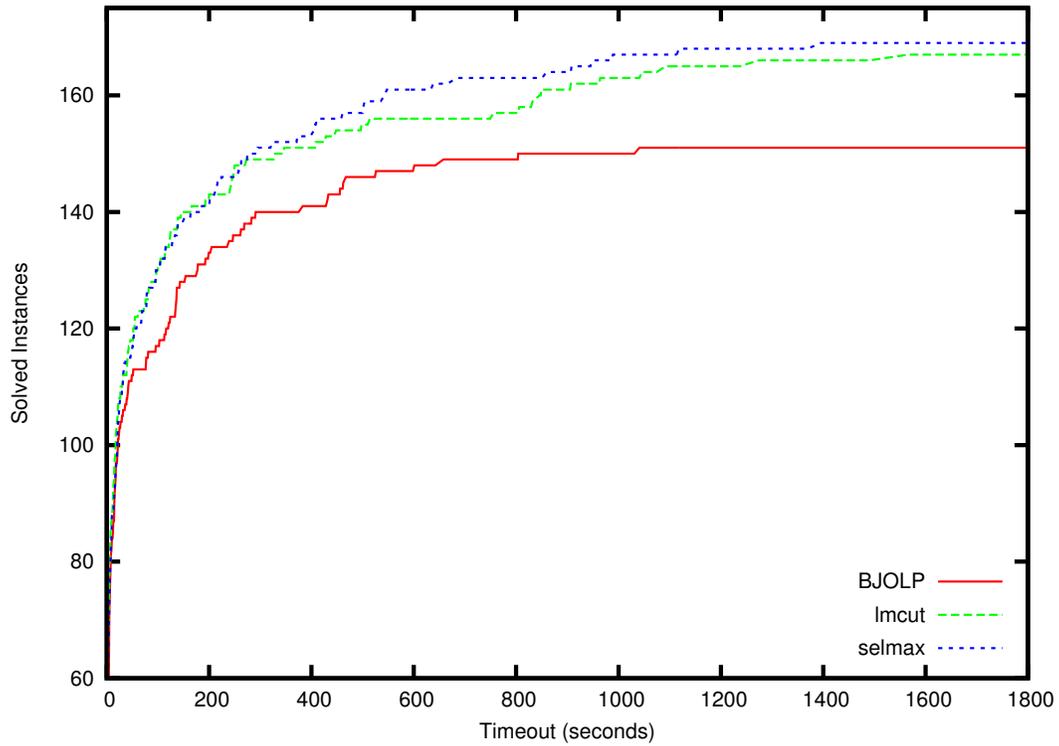

Figure 3: IPC-2011 anytime performance. Each line shows the number of problems from IPC-2011 solved by the *BJOLP*, *lmcut*, and *selmax* planners, respectively, under different timeouts.

| Domain | BJOLP | lmcut | selmax | FDSS-1 |
|---|---|---|---|---|
| barman | **4** | **4** | **4** | 4 |
| elevators | 14 | **18** | **18** | 18 |
| floortile | 2 | **7** | **7** | 7 |
| nomystery | **20** | 15 | **20** | 20 |
| openstacks | 14 | **16** | 14 | 16 |
| parcprinter | 11 | **13** | **13** | 14 |
| parking | 3 | 2 | **4** | 7 |
| pegsol | 17 | **18** | 17 | 19 |
| scanalyzer | 6 | **12** | 10 | 14 |
| sokoban | **20** | **20** | **20** | 20 |
| tidybot | **14** | **14** | **14** | 14 |
| transport | **7** | 6 | 6 | 7 |
| visitall | **10** | **10** | **10** | 13 |
| woodworking | 9 | **12** | **12** | 12 |
| **TOTAL** | 151 | 167 | **169** | 185 |

Table 2: Number of planning tasks solved at IPC 2011 in each domain by the *BJOLP*, *lmcut*, and *selmax* planners. The best result from these 3 planners is in bold. The number of problems solved by Fast Downward Stone Soup 1 (FDSS-1) in each domain is also included for reference.

While there are other admissible heuristics for $\text{SAS}^+$ planning that are competitive with the three above (for example, Helmert, Haslum, & Hoffmann, 2007; Nissim et al., 2011; Katz & Domshlak, 2010), they are based on expensive offline preprocessing, followed by very fast online per-state computation. In contrast, $h_{LA}$, $h_{\text{LM-CUT}}$ and $h_{\text{LM-CUT}^+}$ perform most of their computation online, and thus can be better exploited by selective max.

Additionally, we empirically examine the effectiveness of selective max in deciding whether to compute a heuristic value at all. This is done by combining our most accurate heuristic, $h_{\text{LM-CUT}^+}$, with the blind heuristic.

- *Heuristic difference bias $\alpha$:* The hyper-parameter $\alpha$ controls the tradeoff between computation time and heuristic accuracy. Setting $\alpha = 0$ sets the threshold $\tau$ to 0, forcing the decision rule to always choose the more accurate heuristic. Increasing $\alpha$ increases the threshold, forcing the decision rule to choose the more accurate heuristic $h_2$ only if its value is much higher than that of $h_1$. We evaluate selective max with values for $\alpha$ of 0.1, 0.5, 1, 1.5, 2, 3, 4, and 5.

- *Confidence threshold $\rho$:* The confidence threshold $\rho$ controls the active learning part of selective max. Setting $\rho = 0.5$ turns off active learning completely, because the chosen heuristic always comes with a confidence of at least 0.5. Setting $\rho = 1$ would mean using active learning almost always, essentially reducing selective max to regular point-wise maximization. We evaluate selective max with values for $\rho$ of 0.51, 0.6, 0.7, 0.8, 0.9, and 0.99.

- *Initial sample size $N$:* The initial sample size $N$ is an important parameter, not just because it is used to train the initial classifier before any active learning is done, but also because it is the only source of estimates for branching factor, average action cost, and heuristic computation times. It thus affects the threshold $\tau$: Increasing $N$ increases the accuracy of the initial classifier and of the various aforementioned estimates, but also increases the preprocessing time. We evaluate selective max with values for $N$ of 10, 100, and 1000.

- *Sampling method:* The sampling method used to obtain the initial state-space sample is important in that it affects this initial sample, and thus the accuracy of both the threshold $\tau$ and of the initial classifier. We evaluate selective max with three different sampling methods, all described in Section 5.1: biased probes ($\text{sel}_h^P$), unbiased probes ($\text{sel}_h^{UP}$), and the sampling method of Haslum et al. (2007) ($\text{sel}_h^{PDB}$).

- *Classifier:* The choice of classifier is also very important. The Naive Bayes classifier combines very fast learning and classification ($\text{sel}_h^{NB}$). A more sophisticated variant of Naive Bayes called AODE (Webb et al., 2005) is also considered here ($\text{sel}_h^{AODE}$). AODE is more accurate than Naive Bayes, but has higher classification and learning times, as well as increased memory overhead. Another possible choice is using incremental decision trees (Utgoff et al., 1997), which offer even faster classification, but more expensive learning when the tree structure needs to be changed ($\text{sel}_h^{ITI}$). We also consider kNN classifiers (Cover & Hart, 1967), which offer faster learning than Naive Bayes, but usually more expensive classification, especially as $k$ grows larger ($\text{sel}_h^{kNN}$, for $k = 3, 5$).

---

1. Of course, all three heuristics are computable in polynomial time from the $\text{SAS}^+$ description of the planning task.

| Parameter | Default value | Meaning |
|---|---|---|
| Heuristics | $h_{LA}$ / $h_{\text{LM-CUT}}$ | heuristics used |
| $\alpha$ | 1 | heuristic difference bias |
| $\rho$ | 0.6 | confidence threshold |
| $N$ | 100 | initial sample size |
| Sampling method | *PDB* (Haslum et al., 2007) | state-space sampling method |
| Classifier | Naive Bayes | classifier type |

Table 3: Default parameters for $\text{sel}_h$.

Table 3 describes our default values for each of these independent variables. In each of the subsequent experiments, we vary one of these independent variables, keeping the rest at their default values. In all of these experiments, the search for each planning task instance was limited to 30 minutes[2] and to 3 GB of memory. The search times do not include the time needed for translating the planning task from PDDL to $\text{SAS}^+$ and building some of the Fast Downward data structures, which is common to all planners, and is tangential to the issues considered in our study. The search times do include learning and classification time for selective max.

- *Heuristics*

  We begin by varying the set of heuristics in use. For every possible choice of two or more heuristics out of the uniform action cost partitioning version of $h_{LA}$ (which we simply refer to as $h_{LA}$), $h_{\text{LM-CUT}}$ and $h_{\text{LM-CUT}^+}$, we compare selective max to other methods of heuristic combination, as well as to the individual heuristics. We compare selective max ($\text{sel}_h$) to the regular maximum ($\max_h$), as well as to a planner which chooses which heuristic to compute at each state randomly ($\text{rnd}_h$). As it is not clear whether the random choice should favor the more expensive and accurate heuristic or the cheaper and less accurate one, we simply use a uniform random choice.

  This experiment was conducted on all 31 domains with no conditional effects and axioms (which none of the heuristics we used support) from the International Planning Competitions 1998–2008. Because domains vary in difficulty and in the number of tasks, we normalize the score for each planner in each domain between 0 and 1. Normalizing by the number of problems in the domain is not a good idea, as it is always possible to generate any number of effectively unsolvable problems in each domain, so that the fraction of solved problems will approach zero. Therefore, we normalize the number of problems solved in each domain by the number of problems in that domain that were solved by at least one of our planners. While this measure of normalized coverage has the undesirable property that introducing a

---

2. Each search was given a single core of a 3GHz Intel E8400 CPU machine.

| Heuristic | $h_{LA}$ | $h_{\text{LM-CUT}}$ | $h_{\text{LM-CUT}^+}$ |
|---|---|---|---|
| High variance unit cost | **0.89 (175)** | 0.83 (136) | 0.81 (132) |
| Low variance unit cost | **0.98 (345)** | 0.96 (343) | 0.94 (336) |
| Non-uniform cost | 0.80 (136) | **0.94 (160)** | 0.86 (146) |
| TOTAL | 0.91 (**656**) | **0.92** (639) | 0.89 (614) |

(a) Individual Heuristics

| Heuristics | Domains | $\max_h$ | $\text{rnd}_h$ | $\text{sel}_h$ |
|---|---|---|---|---|
| $h_{LA}$ / $h_{\text{LM-CUT}}$ | High variance unit cost | 0.90 (164) | 0.74 (123) | **0.93 (174)** |
| | Low variance unit cost | 0.97 (345) | 0.95 (342) | **0.97 (346)** |
| | Non-uniform cost | 0.92 (156) | 0.79 (138) | **0.93 (157)** |
| | TOTAL | 0.94 (665) | 0.85 (603) | **0.95 (677)** |
| $h_{LA}$ / $h_{\text{LM-CUT}^+}$ | High variance unit cost | 0.84 (149) | 0.68 (115) | **0.90 (164)** |
| | Low variance unit cost | 0.93 (335) | 0.88 (327) | **0.96 (342)** |
| | Non-uniform cost | 0.85 (144) | 0.71 (122) | **0.86 (145)** |
| | TOTAL | 0.89 (628) | 0.78 (564) | **0.92 (651)** |
| $h_{\text{LM-CUT}}$ / $h_{\text{LM-CUT}^+}$ | High variance unit cost | **0.80 (131)** | 0.75 (122) | 0.80 (130) |
| | Low variance unit cost | 0.94 (336) | 0.93 (335) | **0.97 (344)** |
| | Non-uniform cost | 0.87 (147) | 0.86 (145) | **0.93 (156)** |
| | TOTAL | 0.89 (614) | 0.87 (602) | **0.91 (630)** |
| $h_{LA}$ / $h_{\text{LM-CUT}}$ / $h_{\text{LM-CUT}^+}$ | High variance unit cost | 0.84 (149) | 0.69 (116) | **0.87 (154)** |
| | Low variance unit cost | 0.93 (335) | 0.90 (332) | **0.97 (345)** |
| | Non-uniform cost | 0.85 (144) | 0.75 (130) | **0.89 (150)** |
| | TOTAL | 0.89 (628) | 0.81 (578) | **0.92 (649)** |

(b) Combinations of two or more heuristics

Table 4:  Average normalized coverage, and total coverage in parentheses, broken down by groups of domains with unit cost actions and high variance in coverage, domains with unit cost actions and low variance in coverage, and domains with non-uniform action costs. Table (a) shows the results for $A^*$ with individual heuristics, and table (b) shows the results for the maximum ($\max_h$), random choice ($\text{rnd}_h$), and selective max ($\text{sel}_h$) combinations of the set of heuristics listed in each major row.

new planner could change the normalized coverage of the other planners, we believe that it best reflects performance nonetheless. As an overall performance measure, we list the average normalized coverage score across all domains. Using normalized coverage means that domains have equal weight in the aggregate score. Additionally, we list for each domain the number of problems that were solved by any planner (in parentheses next to the domain name), and for each planner we list the number of problems it solved in parentheses.

Tables 4 and 5 summarize the results of this experiment. We divided the domains in our experiment into 3 sets: domains with non-uniform action costs, domains with unit action costs which exhibited a high variance in the number of problems solved between different planners, and domains with unit action costs which exhibited a low variance in the number of problems solved between different planners. We make this distinction because we conducted the following experiments, which examine the effects of the other parameters of selective max, only on the unit cost action domains which exhibited high variance. Tables 4 and 5 summarize the results for these three sets of domains, as well as for all domains combined. Detailed, per-domain results are relegated to Appendix A.

| Heuristics | Domains | $h_{LA}$ | $h_{\text{LM-CUT}}$ | $h_{\text{LM-CUT}+}$ | $\max_h$ | $\text{rnd}_h$ | $\text{sel}_h$ |
|---|---|---|---|---|---|---|---|
| $h_{LA}$ / $h_{\text{LM-CUT}}$ | High variance unit cost | 3.23 | 2.8 | | **1.0** | 3.88 | 1.46 |
| | Low variance unit cost | 3.48 | 1.14 | | **1.0** | 2.14 | 1.2 |
| | Non-uniform cost | 13.23 | 1.01 | | **1.0** | 3.99 | 1.17 |
| | TOTAL | 4.82 | 1.4 | | **1.0** | 2.93 | 1.25 |
| $h_{LA}$ / $h_{\text{LM-CUT}+}$ | High variance unit cost | 4.01 | | 1.77 | **1.0** | 3.17 | 2.16 |
| | Low variance unit cost | 4.55 | | 1.01 | **1.0** | 2.38 | 1.85 |
| | Non-uniform cost | 13.66 | | **1.0** | **1.0** | 3.85 | 1.72 |
| | TOTAL | 5.85 | | 1.16 | **1.0** | 2.9 | 1.89 |
| $h_{\text{LM-CUT}}$ / $h_{\text{LM-CUT}+}$ | High variance unit cost | | 2.29 | 1.01 | **1.0** | 1.7 | 1.24 |
| | Low variance unit cost | | 1.58 | 1.01 | **1.0** | 1.29 | 1.19 |
| | Non-uniform cost | | 1.32 | 1.03 | **1.0** | 1.18 | 1.16 |
| | TOTAL | | 1.66 | 1.01 | **1.0** | 1.35 | 1.2 |
| $h_{LA}$ / $h_{\text{LM-CUT}}$ / $h_{\text{LM-CUT}+}$ | High variance unit cost | 4.06 | 3.81 | 1.78 | **1.0** | 3.61 | 2.1 |
| | Low variance unit cost | 4.65 | 1.59 | 1.02 | **1.0** | 2.05 | 1.57 |
| | Non-uniform cost | 15.2 | 1.37 | 1.03 | **1.0** | 2.74 | 1.49 |
| | TOTAL | 6.1 | 1.91 | 1.18 | **1.0** | 2.56 | 1.67 |

Table 5: Geometric mean of ratio of expansions relative to $\max_h$, broken down by groups of domains with unit cost actions and high variance in coverage, domains with unit cost actions and low variance in coverage, and domains with non-uniform action costs.

Table 4 lists the normalized coverage score, averaged across all domains, and the total number of problems solved in parentheses. Table 4a lists these for each individual heuristic, and Table 4b for every combination method of every set of two or more heuristics. Table 5 shows how accurate each of these heuristic combination methods is. Since, for a given set of base heuristics, $\max_h$ is the most accurate heuristic possible, the accuracy is evaluated relative to $\max_h$. We evaluate each heuristic's accuracy on each task as the number of states expanded by $A^*$ using that heuristic, divided by the number of states expanded by $A^*$ using $\max_h$. We compute the geometric mean for each domain over the tasks solved by all planners of this "accuracy ratio," and list here the geometric mean over these numbers. Each row lists the results for a combination of two or three heuristics; for combinations of two heuristics, we leave the cell representing the heuristic that is not in the combination empty.

Looking at the results of individual heuristics first, we see that the most accurate heuristic ($h_{\text{LM-CUT}+}$) does not do well overall, while the least accurate heuristic ($h_{LA}$) solved the most tasks in total, and $h_{\text{LM-CUT}}$ wins in terms of normalized coverage. However, when looking at the results for individual domains, we see that the best heuristic to use varies, indicating that combining different heuristics could indeed be of practical value.

We now turn our attention to the empirical results for the combinations of all possible subsets of two or more heuristics. The results clearly demonstrate that when more than one heuristic is used, selective max is always better than regular maximum or random choice, both in terms of normalized coverage and absolute number of problems solved. Furthermore, the poor performance of $\text{rnd}_h$, in both coverage and accuracy, demonstrates that the decision rule and the classifier used in selective max are important to its success, and that computing only one heuristic at each state randomly is insufficient, to say the least.

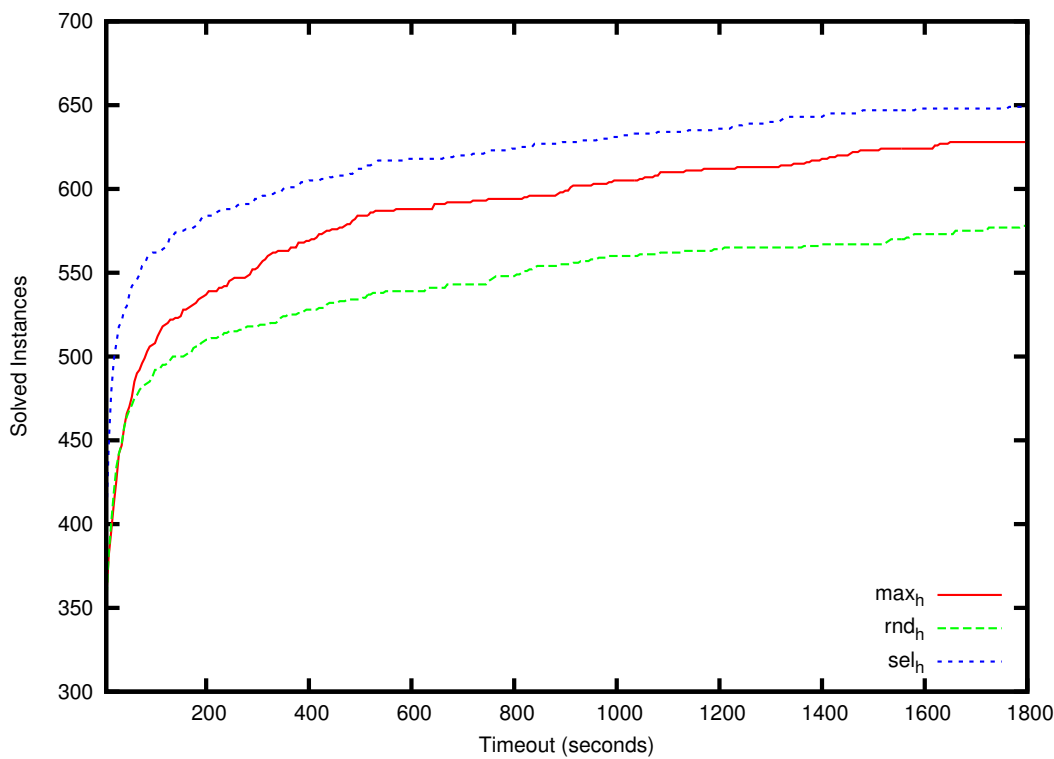

Figure 4: $h_{LA}$ / $h_{\text{LM-CUT}}$ / $h_{\text{LM-CUT}^+}$ anytime profile. Each line shows the number of problems from IPC $1998 - 2006$ solved by the maximum ($\max_h$), random choice ($\text{rnd}_h$), and selective max ($\text{sel}_h$) combination methods of the $h_{LA}$, $h_{\text{LM-CUT}}$, and $h_{\text{LM-CUT}^+}$ heuristics, under different timeouts.

When compared to individual heuristics, selective max does at least as well as each of the individual heuristics it uses, for all combinations except that of $h_{\text{LM-CUT}}$ and $h_{\text{LM-CUT}^+}$. This is most likely because $h_{\text{LM-CUT}}$ and $h_{\text{LM-CUT}^+}$ are based on a very similar procedure, and thus their heuristic estimates are highly correlated. To see why this hinders selective max, consider the extreme case of two heuristics which have a correlation of 1.0 (that is, yield the same heuristic values), where selective max can offer no benefit. Finally, we remark that the best planner in this experiment was the selective max combination of $h_{LA}$ and $h_{\text{LM-CUT}}$.

The above results are all based on a 30 minute time limit, which, while commonly used in the IPC, is arbitrary, and the number of tasks solved after 30 minutes does not tell the complete tale. Here, we examine the anytime profile of the different heuristic combination methods, by plotting the number of tasks solved under different timeouts, up to a timeout of 30 minutes. Figure 4 shows this plot for the three combination methods when all three heuristics are used. As the figure shows, the advantage of $\text{sel}_h$ over the baseline combination methods is even greater under shorter timeouts. This indicates that the advantage of $\text{sel}_h$ over $\max_h$ is even greater than is evident from the results after 30 minutes, and that $\text{sel}_h$ is indeed effective for minimizing search time. Since the anytime plots for the combinations of pairs of heuristics are very similar, we omit them here for the sake of brevity.

| Heuristics | Overhead |
|---|---|
| $h_{LA}$ / $h_{\text{LM-CUT}}$ | 12% |
| $h_{LA}$ / $h_{\text{LM-CUT}^+}$ | 15% |
| $h_{\text{LM-CUT}}$ / $h_{\text{LM-CUT}^+}$ | 9% |
| $h_{LA}$ / $h_{\text{LM-CUT}}$ / $h_{\text{LM-CUT}^+}$ | 10% |

Table 6: Selective max overhead. Each row lists the average percentage of time spent on learning and classification, out of the total time taken by selective max, for each set of heuristics.

Finally, we present overhead statistics for using selective max — the proportion of time spent on learning and classification, including the time spent obtaining the initial state-space sample, out of the total solution time. Table 6 presents the average overhead on selective max for each of the combinations of two or more heuristics. Detailed, per-domain results are presented in Table 18 in Appendix A. As these results show, selective max does incur a noticeable overhead, but it is still relatively low. It is also worth mentioning that the overhead varies significantly between different domains.

We also performed an empirical evaluation of using selective max with an accurate heuristic alongside the blind heuristic. The blind heuristic returns 0 for goal states, and the cost of the cheapest action for non-goal states. For this experiment, we chose our most accurate heuristic, $h_{\text{LM-CUT}^+}$. We compare the performance of $A^*$ using $h_{\text{LM-CUT}^+}$ alone, to that of $A^*$ using selective max of $h_{\text{LM-CUT}^+}$ and the blind heuristic. Because the blind heuristic returns a constant value for all non-goal states, the decision rule that selective max uses to combine some heuristic $h$ with the blind heuristic $h_b$ is simply $h(s) \geq \tau + h_b$, that is, compute $h$ when the predicted value of $h$ is greater than some constant threshold. Recall that, when $h(s) + g(s) < c^*$, computing $h$ is simply a waste of time, because $s$ will not be pruned. Therefore, it only makes sense to compute $h(s)$ when $h(s) \geq c^* - g(s)$. Note that this threshold for computing $h$ depends on $g(s)$, and thus is not constant. This shows that a constant threshold for computing $h(s)$ is not the best possible decision rule. Unfortunately, the selective max decision rule is based on an approximation that fails to capture the subtleties of this case.

Table 7 shows the normalized coverage of $A^*$ using $h_{\text{LM-CUT}^+}$, and $A^*$ using selective max of $h_{\text{LM-CUT}^+}$ and the blind heuristic. As the results show, selective max has little effect in most domains, though it does harm performance in some, and in one domain — OPENSTACKS — it actually performs better than the single heuristic. Table 8 shows the average expansions ratio, using the number of states expanded by $h_{\text{LM-CUT}^+}$ as the baseline; note that using the blind heuristic never increases heuristic accuracy. As these results show, selective max chooses to use the blind heuristic quite often, expanding on average more than twice as many states than $A^*$ with $h_{\text{LM-CUT}^+}$ alone.

| coverage | $h_{\text{LM-CUT}^+}$ | $\text{sel}_h$ |
|---|---|---|
| **airport (31)** | 1.00 (31) | 1.00 (31) |
| **freecell (13)** | 1.00 (13) | 1.00 (13) |
| **logistics00 (17)** | 1.00 (17) | 1.00 (17) |
| **mprime (24)** | 1.00 (24) | 1.00 (24) |
| **mystery (17)** | 1.00 (17) | 1.00 (17) |
| **pipesworld-tankage (9)** | 1.00 (9) | 1.00 (9) |
| **satellite (9)** | 1.00 (9) | 1.00 (9) |
| **zenotravel (12)** | 1.00 (12) | 1.00 (12) |
| **blocks (27)** | 1.00 (27) | 1.00 (27) |
| **depot (7)** | 1.00 (7) | 1.00 (7) |
| **driverlog (14)** | 1.00 (14) | 1.00 (14) |
| **grid (2)** | 1.00 (2) | 1.00 (2) |
| **gripper (6)** | 1.00 (6) | 1.00 (6) |
| **logistics98 (6)** | 1.00 (6) | 1.00 (6) |
| **miconic (140)** | **1.00 (140)** | 0.86 (121) |
| **pathways (5)** | 1.00 (5) | 1.00 (5) |
| **pipesworld-notankage (17)** | 1.00 (17) | 1.00 (17) |
| **psr-small (48)** | 1.00 (48) | 1.00 (48) |
| **rovers (7)** | 1.00 (7) | 1.00 (7) |
| **schedule (27)** | 1.00 (27) | 1.00 (27) |
| **storage (15)** | **1.00 (15)** | 0.93 (14) |
| **tpp (6)** | 1.00 (6) | 1.00 (6) |
| **trucks-strips (9)** | 1.00 (9) | 1.00 (9) |
| **elevators-opt08-strips (18)** | **1.00 (18)** | 0.83 (15) |
| **openstacks-opt08-strips (19)** | 0.89 (17) | **1.00 (19)** |
| **parcprinter-08-strips (21)** | 1.00 (21) | 1.00 (21) |
| **pegsol-08-strips (27)** | 1.00 (27) | 1.00 (27) |
| **scanalyzer-08-strips (13)** | **1.00 (13)** | 0.77 (10) |
| **sokoban-opt08-strips (25)** | 1.00 (25) | 1.00 (25) |
| **transport-opt08-strips (11)** | 1.00 (11) | 1.00 (11) |
| **woodworking-opt08-strips (14)** | **1.00 (14)** | 0.93 (13) |
| **TOTAL** | **1.00 (614)** | 0.98 (589) |

Table 7: Normalized coverage of $h_{\text{LM-CUT}^+}$ and selective max combining $h_{\text{LM-CUT}^+}$ with the blind heuristic. Domains are grouped into domains with unit cost actions and high variance in coverage, domains with unit cost actions and low variance in coverage, and domains with non-uniform action costs, respectively.

| expansions | $h_{\text{LM-CUT}^+}$ | $\text{sel}_h$ |
|---|---|---|
| **airport (31)** | 1.0 | 1.0 |
| **freecell (13)** | **1.0** | 3.13 |
| **logistics00 (17)** | **1.0** | 1.02 |
| **mprime (24)** | **1.0** | 1.22 |
| **mystery (18)** | **1.0** | 3.2 |
| **pipesworld-tankage (9)** | **1.0** | 4.23 |
| **satellite (9)** | **1.0** | 3.11 |
| **zenotravel (12)** | **1.0** | 2.37 |
| **blocks (27)** | **1.0** | 1.92 |
| **depot (7)** | **1.0** | 1.36 |
| **driverlog (14)** | **1.0** | 1.15 |
| **grid (2)** | **1.0** | 7.67 |
| **gripper (6)** | 1.0 | 1.0 |
| **logistics98 (6)** | **1.0** | 1.18 |
| **miconic (121)** | **1.0** | 14.24 |
| **pathways (5)** | 1.0 | 1.0 |
| **pipesworld-notankage (17)** | **1.0** | 1.27 |
| **psr-small (48)** | **1.0** | 2.12 |
| **rovers (7)** | **1.0** | 1.56 |
| **schedule (27)** | **1.0** | 1.21 |
| **storage (14)** | **1.0** | 5.11 |
| **tpp (6)** | **1.0** | 1.6 |
| **trucks-strips (9)** | **1.0** | 1.01 |
| **elevators-opt08-strips (15)** | **1.0** | 13.41 |
| **openstacks-opt08-strips (17)** | **1.0** | 1.08 |
| **parcprinter-08-strips (21)** | **1.0** | 1.24 |
| **pegsol-08-strips (27)** | **1.0** | 1.01 |
| **scanalyzer-08-strips (10)** | **1.0** | 4.87 |
| **sokoban-opt08-strips (25)** | 1.0 | 1.0 |
| **transport-opt08-strips (11)** | **1.0** | 5.86 |
| **woodworking-opt08-strips (13)** | **1.0** | 46.97 |
| **GEOMETRIC MEAN** | **1.0** | 2.3 |

Table 8: Average ratio of expanded states between the baseline of $h_{\text{LM-CUT}^+}$ and selective max combining $h_{\text{LM-CUT}^+}$ with the blind heuristic. Domains are grouped into domains with unit cost actions and high variance in coverage, domains with unit cost actions and low variance in coverage, and domains with non-uniform action costs, respectively.

The above experiments have varied the heuristics which selective max uses. In the following experiments, we fix the set of heuristics, and examine the impact of the other parameters of selective max on performance. As we still need to evaluate over 20 different configurations of selective max, we will focus on eight selected domains: AIRPORT, FREECELL, LOGISTICS00, MPRIME, MYSTERY, PIPESWORLD-TANKAGE, SATELLITE, and ZENOTRAVEL. These are the eight domains with the highest observed variance in the number of tasks solved across different planners, out of the unit action cost domains we used. These domains were chosen in order to reduce the computation time required for these experiments to a manageable quantity. We excluded domains with non-uniform action costs, because they use a different method of estimating the goal depth for the state-space sampling method, which is one of the parameters we examine. Below, we focus on one parameter of selective max at a time, and present the total number of tasks solved in our eight chosen domains, under different values of that parameter. Detailed, per-domain results for each parameter appear in Appendix A.

- *hyper-parameter $\alpha$*

  Figure 5a plots the total number of problems solved, under different values of $\alpha$. As these results show, selective max is fairly robust with respect to the value of $\alpha$, unless a very large value for $\alpha$ is chosen, making it more difficult for selective max to choose the more accurate heuristic.

  Detailed, per-domain results appear in Table 19 in Appendix A, as well as in Figure 6. These results show a more complex picture, where there seems to be some cutoff value for each domain, such that increasing $\alpha$ past that value impairs performance. The one exception to this is the PIPESWORLD-TANKAGE domain, where setting $\alpha = 5$ helps.

- *confidence threshold $\rho$*

  Figure 5b plots the total number of problems solved, under different values of $\rho$, Detailed, per-domain results appear in Table 20 in Appendix A. These results indicate that selective max is also robust to values of $\rho$, unless it is set to a very low value, causing selective max to behave like the regular point-wise maximum.

- *initial sample size $N$*

  Figure 5c plots the total number of problems solved under different values of $N$. with the $x$-axis in logscale. Detailed, per-domain results appear in Table 21 in Appendix A. As the results show, our default value of $N = 100$ is the best (of the three values we tried), although selective max is still fairly robust with respect to the choice of parameter.

- *sampling method*

  Figure 7 shows the total number of problems solved using different methods for the initial state-space sampling. Detailed, per-domain results appear in Table 22 in Appendix A. As the results demonstrate, the choice of sampling method can notably affect the performance of selective max. However, as the detailed results show, this effect is only evident in the FREECELL domain. We also remark that our default sampling method, *PDB*, performs worse than the others. Indeed by using the probe based sampling methods, selective max outperforms $A^*$ using $h_{LA}$ alone. However, as this difference is only due to the FREECELL domain, we can not state with certainty that this would generalize across all domains.

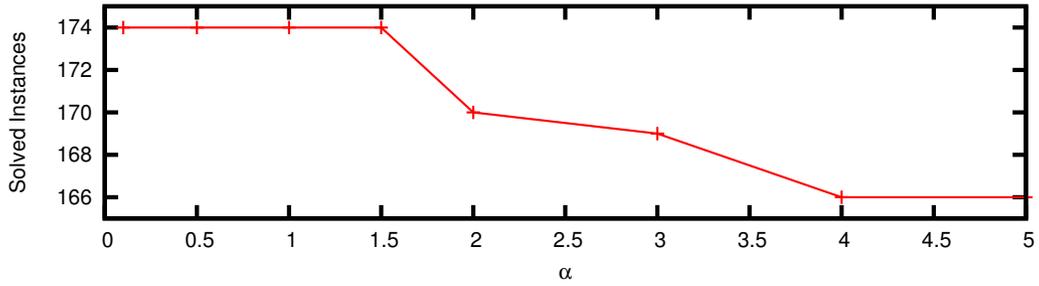

(a) Hyper-parameter $\alpha$

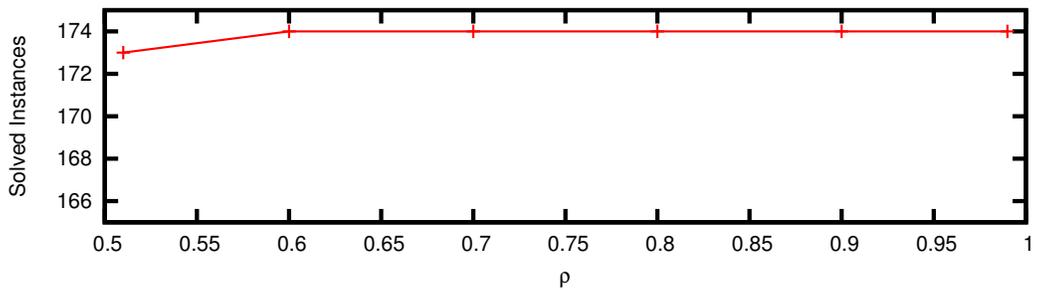

(b) Confidence threshold $\rho$

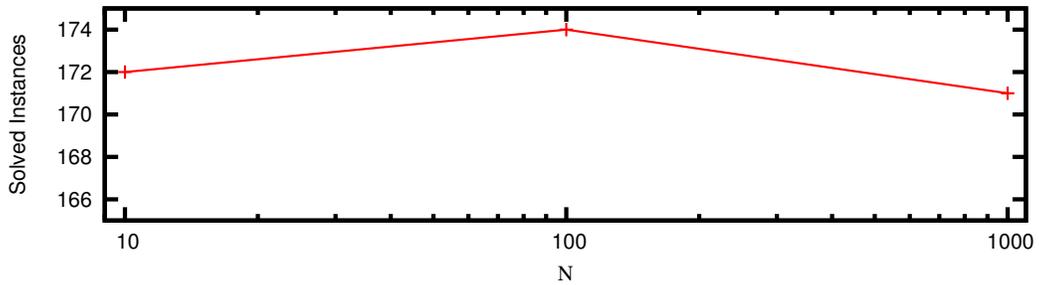

(c) Initial Sample Size $N$

Figure 5: Number of problems solved by selective max under different values for (a) hyper-parameter $\alpha$ (b) confidence threshold $\rho$, and (c) initial sample size $N$.

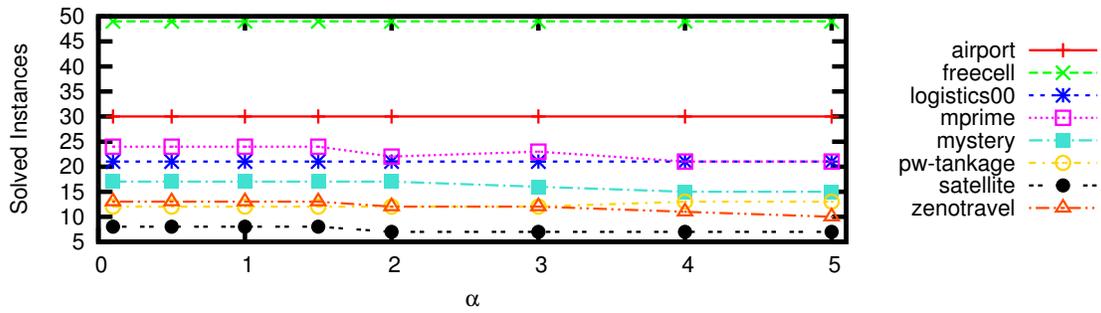

Figure 6: Number of problems solved by selective in each domain under different values for $\alpha$.

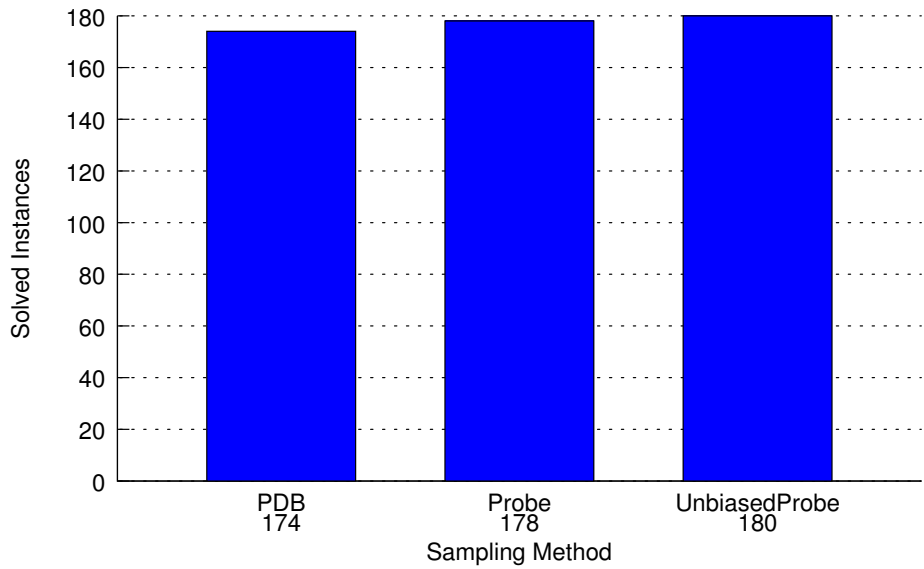

Figure 7: Number of problems solved by selective max with different sampling methods.

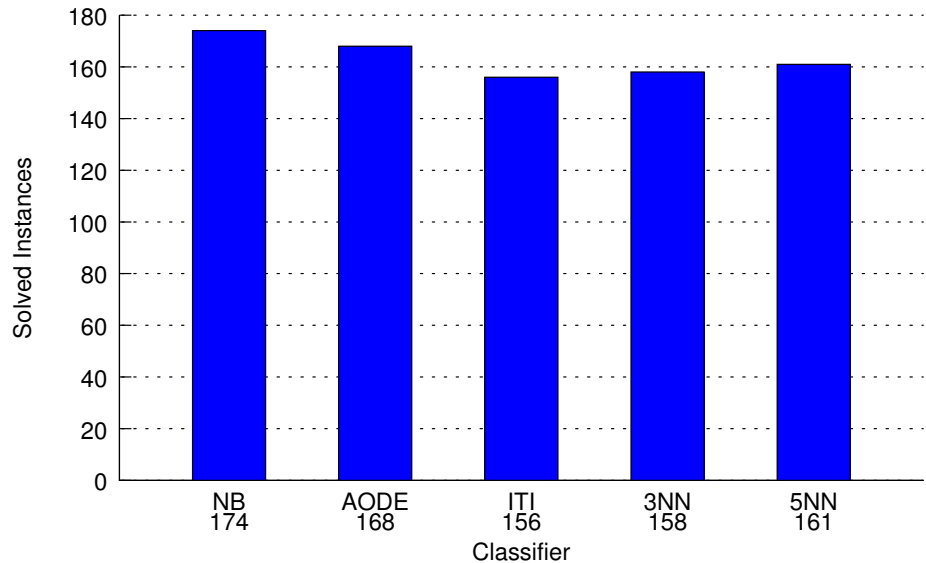

Figure 8: Number of problems solved by selective max with different classifiers.

- *classifier*

  Figure 8 shows the total number of problems solved using different classifiers. Detailed, per-domain results appear in Table 23 in Appendix A. Naive Bayes appears to be the best classifier to use with selective max, although AODE also performs quite well. Even though kNN enjoys very fast learning, the classifier is used mostly for classification, and as expected, kNN does not do well. However, the increased accuracy of $k = 5$ seems to pay off against the faster classification when $k = 3$.

### 6.3 Comparison with Sequential Portfolios

Sequential portfolio solvers for optimal planning are another approach for exploiting the merits of different heuristic functions, and they have been very successful in practice, with the Fast Downward Stone Soup sequential portfolio (Helmert et al., 2011) winning the sequential optimal track at IPC-2011. A sequential portfolio utilizes different solvers by running them sequentially, each with a pre-specified time limit. If one solver fails to find a solution under its allotted time limit, the sequential portfolio terminates it, and moves on to the next solver. However, a sequential portfolio solver needs to know the time allowance for the problem it is trying to solve beforehand, a setting known as *contract anytime* (Russell & Zilberstein, 1991). In contrast, selective max can be used in an *interruptible anytime* manner, where the time limit need not be known in advance.

Here, we compare selective max to sequential portfolios of $A^*$ with the same heuristics. As we have the exact time it took $A^*$ search using each heuristic alone to solve each problem, we can determine whether a sequential portfolio which assigns each heuristic some time limit will be able to solve each problem. Using this data, we simulate the results of two types of sequential portfolio planners. In the first setting, we assume that the time limit is known in advance, and simulate the results of a contract portfolio giving an equal share of time to all heuristics. In the second setting, we simulate an interruptible anytime portfolio by using binary exponential backoff time limits: starting with a time limit of 1 second for each heuristic, we increase the time limit by a factor of 2 if none of the heuristics were able to guide $A^*$ to solve the planning problem. There are several possible orderings for the heuristics here, and we use the *de facto* best ordering for each problem. We denote the contract anytime portfolio by $\mathrm{port}_{ctr}$, and the interruptible anytime portfolio by $\mathrm{port}_{int}$.

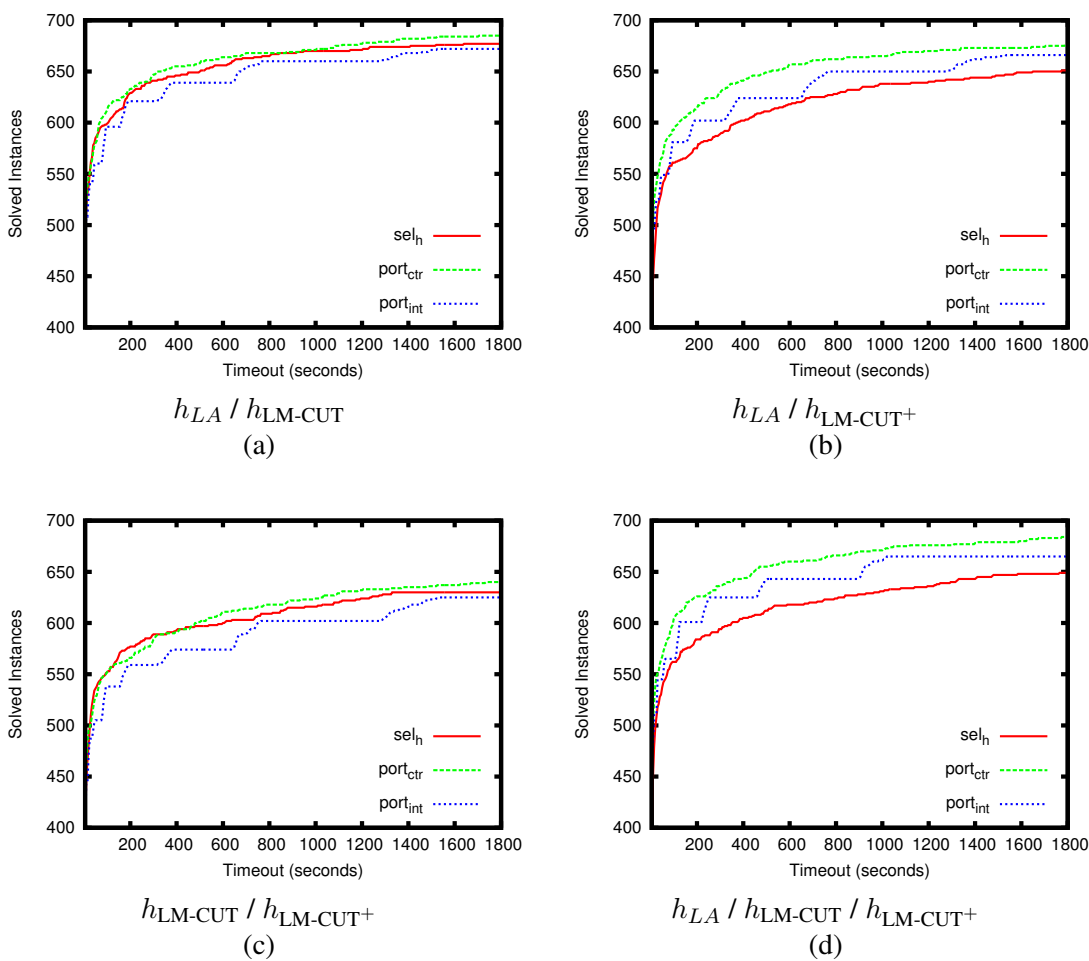

Figure 9: Anytime profiles of sequential portfolios and selective max. Each plot shows the number of problems solved by selective max ($\mathrm{sel}_h$), a simulated contract anytime portfolio ($\mathrm{port}_{ctr}$), and a simulated interruptible portfolio ($\mathrm{port}_{int}$) using (a) $h_{LA}$ and $h_{\text{LM-CUT}}$ (b) $h_{LA}$ and $h_{\text{LM-CUT}^+}$ (c) $h_{\text{LM-CUT}}$ and $h_{\text{LM-CUT}^+}$, and (d) $h_{LA}$, $h_{\text{LM-CUT}}$, and $h_{\text{LM-CUT}^+}$.

Figure 9 shows the number of problems solved under different time limits for selective max, the contract anytime sequential portfolio, and the interruptible anytime sequential portfolio. As these results show, the contract anytime sequential portfolio almost always outperforms selective max. On the other hand, when the sequential portfolio does not know the time limit in advance, its performance deteriorates significantly. The best heuristic combination for selective max, $h_{LA}$ and $h_{\text{LM-CUT}}$, outperforms the interruptible anytime portfolio using the same heuristics, and so does the selective max combination of $h_{\text{LM-CUT}}$ and $h_{\text{LM-CUT}^+}$. With the other combinations of heuristics, the interruptible anytime portfolio performs better than selective max.

## 7. Discussion

Learning for planning has been a very active field since the early days of planning (Fikes, Hart, & Nilsson, 1972), and is recently receiving growing attention in the community. However, despite some early work (Rendell, 1983), relatively little work has dealt with learning for state-space search guided by distance-estimating heuristics, one of the most prominent approaches to planning these days. Most works in this direction have been devoted to learning macro-actions (see, for example, Finkelstein & Markovitch, 1998; Botea, Enzenberger, Müller, & Schaeffer, 2005; Coles & Smith, 2007). Recently, learning for heuristic search planning has received more attention: Yoon et al. (2008) suggested learning (inadmissible) heuristic functions based upon features extracted from relaxed plans. Arfaee, Zilles, and Holte (2010) attempted to learn an almost admissible heuristic estimate using a neural network. Perhaps the most closely related work to ours is that of Thayer, Dionne, and Ruml (2011), who learn to correct errors in heuristic estimates online. Thayer et al. attempt to improve the accuracy of a single given heuristic, while selective max attempts to choose one of several given heuristics for each state. The two works differ technically on this point. More importantly, however, none of the aforementioned approaches can guarantee that the resulting heuristic will be admissible, and thus that an optimal solution will be found. In contrast, our focus is on optimal planning, and we are not aware of any previous work that deals with learning for optimal heuristic search.

Our experimental evaluation demonstrates that selective max is a more effective method for combining arbitrary admissible heuristics than the baseline point-wise maximization. Also advantageous is selective max's ability to exploit pairs of heuristics, where one is guaranteed to always be at least as accurate as the other. For example, the $h_{LA}$ heuristic can be used with two action cost partitioning schemes: uniform and optimal (Karpas & Domshlak, 2009). The heuristic induced by the optimal action cost partitioning is at least as accurate the one induced by the uniform action cost partitioning, but takes much longer to compute. Selective max might be used to learn when it is worth spending the extra time to compute the optimal cost partitioning, and when it is not. In contrast, the max-based combination of these two heuristics would simply waste the time spent on computing the uniform action cost partitioning.

The controlled parametric experiments demonstrate that the right choice of classifier and of the sampling method for the initial state-space sample is very important. The other parameters of selective max do not appear to affect performance too much, as long as they are set to reasonable values. This implies that selective max could be improved by using faster, more accurate, classifiers, and by developing sampling methods that can represent the state-space well.

Finally, we remark that the Fast Downward Autotune entry in the sequential optimal track of the 2011 edition of the International Planning Competition, which used ParamILS (Hutter, Hoos, Leyton-Brown, & Stützle, 2009) to choose the "best" configuration for the Fast Downward planner, chose to use selective-max to combine $h_{\text{LM-CUT}}$ and $h_{max}$ (Bonet, Loerincs, & Geffner, 1997). This provides further evidence that selective max is a practically valuable method for combining heuristics in optimal planning.

| coverage | $h_{LA}$ | $h_{\text{LM-CUT}}$ | $h_{\text{LM-CUT}}+$ |
|---|---|---|---|
| airport (33) | 0.91 (30) | 0.85 (28) | **0.94 (31)** |
| freecell (58) | **1.00 (58)** | 0.26 (15) | 0.22 (13) |
| logistics00 (21) | **1.00 (21)** | 0.95 (20) | 0.81 (17) |
| mprime (24) | 0.88 (21) | **1.00 (24)** | **1.00 (24)** |
| mystery (17) | 0.88 (15) | **1.00 (17)** | **1.00 (17)** |
| pipesworld-tankage (13) | **1.00 (13)** | 0.92 (12) | 0.69 (9) |
| satellite (10) | 0.70 (7) | 0.70 (7) | **0.90 (9)** |
| zenotravel (13) | 0.77 (10) | **1.00 (13)** | 0.92 (12) |
| blocks (28) | 0.96 (27) | **1.00 (28)** | 0.96 (27) |
| depot (7) | 1.00 (7) | 1.00 (7) | 1.00 (7) |
| driverlog (14) | **1.00 (14)** | 0.93 (13) | **1.00 (14)** |
| grid (3) | **1.00 (3)** | 0.67 (2) | 0.67 (2) |
| gripper (7) | **1.00 (7)** | **1.00 (7)** | 0.86 (6) |
| logistics98 (6) | 1.00 (6) | 1.00 (6) | 1.00 (6) |
| miconic (142) | **1.00 (142)** | 0.99 (141) | 0.99 (140) |
| pathways (5) | 0.80 (4) | **1.00 (5)** | **1.00 (5)** |
| pipesworld-notankage (18) | **1.00 (18)** | 0.94 (17) | 0.94 (17) |
| psr-small (49) | **1.00 (49)** | **1.00 (49)** | 0.98 (48) |
| rovers (8) | **1.00 (8)** | 0.88 (7) | 0.88 (7) |
| schedule (30) | **1.00 (30)** | **1.00 (30)** | 0.90 (27) |
| storage (15) | 1.00 (15) | 1.00 (15) | 1.00 (15) |
| tpp (6) | 1.00 (6) | 1.00 (6) | 1.00 (6) |
| trucks-strips (10) | 0.90 (9) | **1.00 (10)** | 0.90 (9) |
| elevators-opt08-strips (22) | 0.77 (17) | **1.00 (22)** | 0.82 (18) |
| openstacks-opt08-strips (20) | 0.90 (18) | **1.00 (20)** | 0.85 (17) |
| parcprinter-08-strips (22) | 0.68 (15) | 0.82 (18) | **0.95 (21)** |
| pegsol-08-strips (28) | 0.96 (27) | **1.00 (28)** | 0.96 (27) |
| scanalyzer-08-strips (16) | 0.56 (9) | **0.94 (15)** | 0.81 (13) |
| sokoban-opt08-strips (30) | 0.83 (25) | **1.00 (30)** | 0.83 (25) |
| transport-opt08-strips (12) | **1.00 (12)** | 0.92 (11) | 0.92 (11) |
| woodworking-opt08-strips (19) | 0.68 (13) | **0.84 (16)** | 0.74 (14) |
| TOTAL | 0.91 (656) | **0.92 (639)** | 0.89 (614) |

Table 9: Detailed per-domain results of $A^*$ with each individual heuristic. Normalized coverage is shown, with the number of problems solved shown in parentheses. Domains are grouped into domains with unit cost actions and high variance in coverage, domains with unit cost actions and low variance in coverage, and domains with non-uniform action costs, respectively.

## Acknowledgments

The work was partly supported by the Israel Science Foundation (ISF) grant 1045/12.

## Appendix A. Detailed Results of Empirical Evaluation

In this appendix, we present detailed per-domain, results of the experiments described in Section 6.

Table 9 shows the normalized coverage and number of problems solved in each domain, for individual heuristics. The normalized coverage score of planner $X$ on domain $D$ is the number of problems from domain $D$ solved by planner $X$, divided by the number of problems from domain $D$ solved by at least one planner. Tables 10 – 17 give the results for combinations of two or more heuristics. Tables 10, 12, 14, and 16 list the normalized coverage of the individual heuristics used, and of their combination using selective max ($\text{sel}_h$), regular maximum ($\text{max}_h$), and random choice of heuristic at each state ($\text{rnd}_h$) after 30 minutes. Tables 11, 13, 15, and 17 give the geometric mean of the ratio of expanded states relative to $\text{max}_h$ in each domain, over problems solved by all configurations. The number of tasks solved by all planners is listed in parentheses next to each domain. The final row gives the geometric mean over the geometric means of each domain.

| coverage | $h_{LA}$ | $h_{\textbf{LM-CUT}}$ | $\max_h$ | $\text{rnd}_h$ | $\text{sel}_h$ |
|---|---|---|---|---|---|
| **airport (33)** | **0.91 (30)** | 0.85 (28) | **0.91 (30)** | 0.85 (28) | **0.91 (30)** |
| **freecell (58)** | **1.00 (58)** | 0.26 (15) | 0.71 (41) | 0.28 (16) | 0.84 (49) |
| **logistics00 (21)** | **1.00 (21)** | 0.95 (20) | 0.95 (20) | 0.95 (20) | **1.00 (21)** |
| **mprime (24)** | 0.88 (21) | **1.00 (24)** | **1.00 (24)** | 0.75 (18) | **1.00 (24)** |
| **mystery (17)** | 0.88 (15) | **1.00 (17)** | **1.00 (17)** | 0.76 (13) | **1.00 (17)** |
| **pipesworld-tankage (13)** | **1.00 (13)** | 0.92 (12) | 0.92 (12) | 0.85 (11) | 0.92 (12) |
| **satellite (10)** | 0.70 (7) | 0.70 (7) | 0.70 (7) | 0.70 (7) | **0.80 (8)** |
| **zenotravel (13)** | 0.77 (10) | **1.00 (13)** | **1.00 (13)** | 0.77 (10) | **1.00 (13)** |
| **blocks (28)** | 0.96 (27) | **1.00 (28)** | **1.00 (28)** | **1.00 (28)** | **1.00 (28)** |
| **depot (7)** | **1.00 (7)** | 1.00 (7) | 1.00 (7) | 1.00 (7) | 1.00 (7) |
| **driverlog (14)** | **1.00 (14)** | 0.93 (13) | **1.00 (14)** | 0.93 (13) | **1.00 (14)** |
| **grid (3)** | **1.00 (3)** | 0.67 (2) | 0.67 (2) | 0.67 (2) | 0.67 (2) |
| **gripper (7)** | 1.00 (7) | 1.00 (7) | 1.00 (7) | 1.00 (7) | 1.00 (7) |
| **logistics98 (6)** | 1.00 (6) | 1.00 (6) | 1.00 (6) | 1.00 (6) | 1.00 (6) |
| **miconic (142)** | **1.00 (142)** | 0.99 (141) | 0.99 (141) | 0.99 (141) | **1.00 (142)** |
| **pathways (5)** | 0.80 (4) | **1.00 (5)** | **1.00 (5)** | 0.80 (4) | **1.00 (5)** |
| **pipesworld-notankage (18)** | **1.00 (18)** | 0.94 (17) | 0.94 (17) | 0.94 (17) | 0.94 (17) |
| **psr-small (49)** | 1.00 (49) | 1.00 (49) | 1.00 (49) | 1.00 (49) | 1.00 (49) |
| **rovers (8)** | **1.00 (8)** | 0.88 (7) | **1.00 (8)** | **1.00 (8)** | **1.00 (8)** |
| **schedule (30)** | 1.00 (30) | 1.00 (30) | 1.00 (30) | 1.00 (30) | 1.00 (30) |
| **storage (15)** | 1.00 (15) | 1.00 (15) | 1.00 (15) | 1.00 (15) | 1.00 (15) |
| **tpp (6)** | 1.00 (6) | 1.00 (6) | 1.00 (6) | 1.00 (6) | 1.00 (6) |
| **trucks-strips (10)** | 0.90 (9) | **1.00 (10)** | **1.00 (10)** | 0.90 (9) | **1.00 (10)** |
| **elevators-opt08-strips (22)** | 0.77 (17) | **1.00 (22)** | **1.00 (22)** | 0.77 (17) | **1.00 (22)** |
| **openstacks-opt08-strips (20)** | 0.90 (18) | **1.00 (20)** | 0.90 (18) | 0.90 (18) | 0.90 (18) |
| **parcprinter-08-strips (22)** | 0.68 (15) | **0.82 (18)** | **0.82 (18)** | 0.68 (15) | **0.82 (18)** |
| **pegsol-08-strips (28)** | 0.96 (27) | **1.00 (28)** | 0.96 (27) | 0.96 (27) | 0.96 (27) |
| **scanalyzer-08-strips (16)** | 0.56 (9) | **0.94 (15)** | **0.94 (15)** | 0.44 (7) | **0.94 (15)** |
| **sokoban-opt08-strips (30)** | 0.83 (25) | **1.00 (30)** | 0.97 (29) | **1.00 (30)** | 0.97 (29) |
| **transport-opt08-strips (12)** | **1.00 (12)** | 0.92 (11) | 0.92 (11) | 0.92 (11) | 0.92 (11) |
| **woodworking-opt08-strips (19)** | 0.68 (13) | 0.84 (16) | 0.84 (16) | 0.68 (13) | **0.89 (17)** |
| **TOTAL** | 0.91 (656) | 0.92 (639) | 0.94 (665) | 0.85 (603) | **0.95 (677)** |

Table 10: Detailed per-domain normalized coverage using $h_{LA}$ and $h_{\text{LM-CUT}}$. Each line shows the normalized coverage in each domain, with the number of problems solved is shown in parentheses. Domains are grouped into domains with unit cost actions and high variance in coverage, domains with unit cost actions and low variance in coverage, and domains with non-uniform action costs, respectively.

| expansions | $h_{LA}$ | $h_{\text{LM-CUT}}$ | $\max_h$ | $\text{rnd}_h$ | $\text{sel}_h$ |
|---|---|---|---|---|---|
| **airport (28)** | 2.88 | 1.12 | **1.0** | 1.61 | 2.2 |
| **freecell (15)** | 1.01 | 529.61 | **1.0** | 116.96 | 2.14 |
| **logistics00 (20)** | 1.0 | 1.0 | 1.0 | 1.0 | 1.0 |
| **mprime (18)** | 6.34 | 1.89 | **1.0** | 4.2 | 1.52 |
| **mystery (14)** | 7.9 | 1.15 | **1.0** | 5.19 | 1.17 |
| **pipesworld-tankage (11)** | 1.61 | 2.35 | **1.0** | 1.62 | 1.12 |
| **satellite (7)** | 6.27 | 1.26 | **1.0** | 2.32 | 1.09 |
| **zenotravel (10)** | 7.98 | 1.0 | **1.0** | 3.3 | 2.02 |
| **blocks (27)** | 7.4 | **1.0** | **1.0** | 2.2 | 1.61 |
| **depot (7)** | 3.45 | 1.32 | **1.0** | 1.91 | 1.3 |
| **driverlog (13)** | 7.2 | 1.09 | **1.0** | 2.89 | 1.24 |
| **grid (2)** | 2.15 | 1.73 | **1.0** | 1.57 | 1.83 |
| **gripper (7)** | **1.0** | 1.04 | **1.0** | 1.02 | **1.0** |
| **logistics98 (6)** | 7.74 | **1.0** | **1.0** | 2.69 | 1.08 |
| **miconic (141)** | 1.0 | 1.0 | 1.0 | 1.0 | 1.0 |
| **pathways (4)** | 39.65 | **1.0** | **1.0** | 17.91 | **1.0** |
| **pipesworld-notankage (17)** | 2.01 | 2.16 | **1.0** | 1.97 | 1.36 |
| **psr-small (49)** | 1.27 | **1.0** | **1.0** | 1.11 | 1.15 |
| **rovers (7)** | 2.18 | 1.31 | **1.0** | 1.77 | 1.09 |
| **schedule (30)** | 1.15 | 1.0 | **1.0** | 1.03 | 1.15 |
| **storage (15)** | 2.16 | **1.0** | **1.0** | 1.45 | 1.56 |
| **tpp (6)** | 1.74 | **1.0** | **1.0** | 1.42 | **1.0** |
| **trucks-strips (9)** | 46.11 | 1.02 | **1.0** | 12.12 | 1.01 |
| **elevators-opt08-strips (17)** | 21.51 | 1.03 | **1.0** | 5.99 | 1.37 |
| **openstacks-opt08-strips (18)** | 1.17 | **1.0** | **1.0** | 1.03 | 1.15 |
| **parcprinter-08-strips (15)** | 24.13 | **1.0** | **1.0** | 9.34 | **1.0** |
| **pegsol-08-strips (27)** | 3.72 | 1.01 | **1.0** | 1.8 | 1.01 |
| **scanalyzer-08-strips (7)** | 69.2 | **1.0** | **1.0** | 21.47 | 1.14 |
| **sokoban-opt08-strips (25)** | 15.74 | 1.07 | **1.0** | 1.33 | 1.04 |
| **transport-opt08-strips (11)** | 12.09 | 1.01 | **1.0** | 3.79 | 1.44 |
| **woodworking-opt08-strips (12)** | 31.6 | **1.0** | **1.0** | 5.68 | 1.28 |
| **GEOMETRIC MEAN** | 4.82 | 1.4 | **1.0** | 2.93 | 1.25 |

Table 11: Detailed per-domain expansions relative to $\max_h$ using $h_{LA}$ and $h_{\text{LM-CUT}}$. Each row shows the geometric mean of the ratio of expanded nodes relative to $\max_h$. Domains are grouped into domains with unit cost actions and high variance in coverage, domains with unit cost actions and low variance in coverage, and domains with non-uniform action costs, respectively.

| coverage | $h_{LA}$ | $h_{\text{LM-CUT}^+}$ | $\text{max}_h$ | $\text{rnd}_h$ | $\text{sel}_h$ |
|---|---|---|---|---|---|
| **airport (33)** | 0.91 (30) | **0.94 (31)** | **0.94 (31)** | 0.85 (28) | 0.91 (30) |
| **freecell (58)** | **1.00 (58)** | 0.22 (13) | 0.53 (31) | 0.26 (15) | 0.71 (41) |
| **logistics00 (21)** | **1.00 (21)** | 0.81 (17) | 0.76 (16) | 0.95 (20) | **1.00 (21)** |
| **mprime (24)** | 0.88 (21) | **1.00 (24)** | **1.00 (24)** | 0.67 (16) | **1.00 (24)** |
| **mystery (17)** | 0.88 (15) | **1.00 (17)** | **1.00 (17)** | 0.71 (12) | **1.00 (17)** |
| **pipesworld-tankage (13)** | **1.00 (13)** | 0.69 (9) | 0.69 (9) | 0.62 (8) | 0.69 (9) |
| **satellite (10)** | 0.70 (7) | 0.90 (9) | 0.90 (9) | 0.70 (7) | **1.00 (10)** |
| **zenotravel (13)** | 0.77 (10) | **0.92 (12)** | **0.92 (12)** | 0.69 (9) | **0.92 (12)** |
| **blocks (28)** | **0.96 (27)** | **0.96 (27)** | **0.96 (27)** | 0.93 (26) | 0.93 (26) |
| **depot (7)** | **1.00 (7)** | **1.00 (7)** | **1.00 (7)** | 0.86 (6) | **1.00 (7)** |
| **driverlog (14)** | **1.00 (14)** | **1.00 (14)** | **1.00 (14)** | 0.93 (13) | 0.93 (13) |
| **grid (3)** | **1.00 (3)** | 0.67 (2) | 0.67 (2) | 0.67 (2) | 0.67 (2) |
| **gripper (7)** | **1.00 (7)** | 0.86 (6) | 0.71 (5) | 0.86 (6) | **1.00 (7)** |
| **logistics98 (6)** | **1.00 (6)** | **1.00 (6)** | **1.00 (6)** | 0.83 (5) | **1.00 (6)** |
| **miconic (142)** | **1.00 (142)** | 0.99 (140) | 0.99 (140) | 0.99 (140) | **1.00 (142)** |
| **pathways (5)** | 0.80 (4) | **1.00 (5)** | **1.00 (5)** | 0.80 (4) | **1.00 (5)** |
| **pipesworld-notankage (18)** | **1.00 (18)** | 0.94 (17) | 0.94 (17) | 0.83 (15) | 0.94 (17) |
| **psr-small (49)** | **1.00 (49)** | 0.98 (48) | 0.98 (48) | 0.98 (48) | **1.00 (49)** |
| **rovers (8)** | **1.00 (8)** | 0.88 (7) | 0.88 (7) | 0.88 (7) | **1.00 (8)** |
| **schedule (30)** | **1.00 (30)** | 0.90 (27) | 0.90 (27) | 0.90 (27) | **1.00 (30)** |
| **storage (15)** | 1.00 (15) | 1.00 (15) | 1.00 (15) | 1.00 (15) | 1.00 (15) |
| **tpp (6)** | 1.00 (6) | 1.00 (6) | 1.00 (6) | 1.00 (6) | 1.00 (6) |
| **trucks-strips (10)** | **0.90 (9)** | **0.90 (9)** | **0.90 (9)** | 0.70 (7) | **0.90 (9)** |
| **elevators-opt08-strips (22)** | 0.77 (17) | **0.82 (18)** | **0.82 (18)** | 0.59 (13) | 0.73 (16) |
| **openstacks-opt08-strips (20)** | **0.90 (18)** | 0.85 (17) | 0.80 (16) | 0.85 (17) | 0.85 (17) |
| **parcprinter-08-strips (22)** | 0.68 (15) | 0.95 (21) | 0.95 (21) | 0.55 (12) | **1.00 (22)** |
| **pegsol-08-strips (28)** | 0.96 (27) | 0.96 (27) | 0.96 (27) | 0.96 (27) | 0.96 (27) |
| **scanalyzer-08-strips (16)** | 0.56 (9) | **0.81 (13)** | **0.81 (13)** | 0.38 (6) | **0.81 (13)** |
| **sokoban-opt08-strips (30)** | **0.83 (25)** | **0.83 (25)** | 0.77 (23) | **0.83 (25)** | 0.80 (24) |
| **transport-opt08-strips (12)** | **1.00 (12)** | 0.92 (11) | 0.92 (11) | 0.92 (11) | 0.92 (11) |
| **woodworking-opt08-strips (19)** | 0.68 (13) | 0.74 (14) | **0.79 (15)** | 0.58 (11) | **0.79 (15)** |
| **TOTAL** | 0.91 (656) | 0.89 (614) | 0.89 (628) | 0.78 (564) | **0.92 (651)** |

Table 12: Detailed per-domain normalized coverage using $h_{LA}$ and $h_{\text{LM-CUT}^+}$. Each line shows the normalized coverage in each domain, with the number of problems solved is shown in parentheses. Domains are grouped into domains with unit cost actions and high variance in coverage, domains with unit cost actions and low variance in coverage, and domains with non-uniform action costs, respectively.

| expansions | $h_{LA}$ | $h_{\mathbf{LM\text{-}CUT}^+}$ | $\max_h$ | $\mathrm{rnd}_h$ | $\mathrm{sel}_h$ |
|---|---|---|---|---|---|
| **airport (28)** | 3.05 | **1.0** | **1.0** | 1.43 | 2.81 |
| **freecell (13)** | 1.22 | 47.57 | **1.0** | 10.54 | 2.05 |
| **logistics00 (16)** | 1.0 | 1.0 | 1.0 | 1.0 | 1.0 |
| **mprime (16)** | 8.45 | 1.23 | **1.0** | 5.2 | 1.57 |
| **mystery (13)** | 7.76 | 1.11 | **1.0** | 4.77 | 1.7 |
| **pipesworld-tankage (8)** | 2.17 | 1.42 | **1.0** | 1.48 | 1.86 |
| **satellite (7)** | 19.26 | 1.03 | **1.0** | 5.94 | 4.12 |
| **zenotravel (9)** | 6.62 | **1.0** | **1.0** | 3.09 | 4.04 |
| **blocks (26)** | 6.97 | **1.0** | **1.0** | 2.15 | 4.28 |
| **depot (6)** | 21.8 | **1.0** | **1.0** | 5.46 | 3.96 |
| **driverlog (13)** | 11.11 | 1.01 | **1.0** | 3.71 | 2.56 |
| **grid (2)** | 5.04 | 1.01 | **1.0** | 2.14 | 4.74 |
| **gripper (5)** | 1.0 | 1.0 | 1.0 | 1.0 | 1.0 |
| **logistics98 (5)** | 6.1 | **1.0** | **1.0** | 2.14 | 3.79 |
| **miconic (140)** | 1.0 | 1.0 | 1.0 | 1.0 | 1.0 |
| **pathways (4)** | 40.56 | 1.0 | **1.0** | 18.03 | **1.0** |
| **pipesworld-notankage (15)** | 3.08 | 1.12 | **1.0** | 1.75 | 2.46 |
| **psr-small (48)** | 1.31 | 1.08 | **1.0** | 1.14 | 1.27 |
| **rovers (7)** | 2.75 | 1.01 | **1.0** | 1.81 | 1.45 |
| **schedule (27)** | 1.09 | 1.0 | **1.0** | **1.0** | 1.09 |
| **storage (15)** | 2.29 | **1.0** | **1.0** | 1.53 | 2.16 |
| **tpp (6)** | 2.72 | **1.0** | **1.0** | 1.88 | 1.17 |
| **trucks-strips (7)** | 46.09 | 1.01 | **1.0** | 12.02 | 1.01 |
| **elevators-opt08-strips (13)** | 28.6 | 1.01 | **1.0** | 7.1 | 7.46 |
| **openstacks-opt08-strips (16)** | 1.17 | 1.0 | **1.0** | 1.03 | 1.09 |
| **parcprinter-08-strips (12)** | 24.87 | **1.0** | **1.0** | 9.23 | 1.19 |
| **pegsol-08-strips (27)** | 4.92 | 1.0 | **1.0** | 2.15 | **1.0** |
| **scanalyzer-08-strips (6)** | 23.07 | **1.0** | **1.0** | 6.88 | 1.43 |
| **sokoban-opt08-strips (21)** | 15.66 | 1.0 | **1.0** | 1.33 | 1.01 |
| **transport-opt08-strips (11)** | 15.34 | 1.0 | **1.0** | 4.26 | 2.84 |
| **woodworking-opt08-strips (11)** | 53.27 | **1.0** | **1.0** | 8.53 | 1.91 |
| **GEOMETRIC MEAN** | 5.85 | 1.16 | **1.0** | 2.9 | 1.89 |

Table 13: Detailed per-domain expansions relative to $\max_h$ using $h_{LA}$ and $h_{\mathbf{LM\text{-}CUT}^+}$. Each row shows the geometric mean of the ratio of expanded nodes relative to $\max_h$. Domains are grouped into domains with unit cost actions and high variance in coverage, domains with unit cost actions and low variance in coverage, and domains with non-uniform action costs, respectively.

| coverage | $h_{\text{LM-CUT}}$ | $h_{\text{LM-CUT}^+}$ | $\max_h$ | $\text{rnd}_h$ | $\text{sel}_h$ |
|---|---|---|---|---|---|
| **airport (33)** | 0.85 (28) | **0.94 (31)** | **0.94 (31)** | 0.82 (27) | 0.85 (28) |
| **freecell (58)** | **0.26 (15)** | 0.22 (13) | 0.22 (13) | 0.21 (12) | 0.22 (13) |
| **logistics00 (21)** | **0.95 (20)** | 0.81 (17) | 0.76 (16) | **0.95 (20)** | **0.95 (20)** |
| **mprime (24)** | **1.00 (24)** | **1.00 (24)** | **1.00 (24)** | 0.88 (21) | **1.00 (24)** |
| **mystery (17)** | **1.00 (17)** | **1.00 (17)** | **1.00 (17)** | 0.88 (15) | 0.94 (16) |
| **pipesworld-tankage (13)** | **0.92 (12)** | 0.69 (9) | 0.69 (9) | 0.62 (8) | 0.69 (9) |
| **satellite (10)** | 0.70 (7) | **0.90 (9)** | **0.90 (9)** | 0.70 (7) | 0.80 (8) |
| **zenotravel (13)** | **1.00 (13)** | 0.92 (12) | 0.92 (12) | 0.92 (12) | 0.92 (12) |
| **blocks (28)** | **1.00 (28)** | 0.96 (27) | 0.96 (27) | **1.00 (28)** | **1.00 (28)** |
| **depot (7)** | 1.00 (7) | 1.00 (7) | 1.00 (7) | 1.00 (7) | 1.00 (7) |
| **driverlog (14)** | 0.93 (13) | **1.00 (14)** | **1.00 (14)** | 0.93 (13) | **1.00 (14)** |
| **grid (3)** | 0.67 (2) | 0.67 (2) | 0.67 (2) | 0.67 (2) | 0.67 (2) |
| **gripper (7)** | **1.00 (7)** | 0.86 (6) | 0.86 (6) | 0.86 (6) | **1.00 (7)** |
| **logistics98 (6)** | 1.00 (6) | 1.00 (6) | 1.00 (6) | 1.00 (6) | 1.00 (6) |
| **miconic (142)** | **0.99 (141)** | 0.99 (140) | 0.99 (140) | 0.99 (140) | **0.99 (141)** |
| **pathways (5)** | 1.00 (5) | 1.00 (5) | 1.00 (5) | 1.00 (5) | 1.00 (5) |
| **pipesworld-notankage (18)** | **0.94 (17)** | **0.94 (17)** | **0.94 (17)** | 0.89 (16) | **0.94 (17)** |
| **psr-small (49)** | **1.00 (49)** | 0.98 (48) | 0.98 (48) | 0.98 (48) | **1.00 (49)** |
| **rovers (8)** | 0.88 (7) | 0.88 (7) | 0.88 (7) | 0.88 (7) | 0.88 (7) |
| **schedule (30)** | **1.00 (30)** | 0.90 (27) | 0.90 (27) | 0.90 (27) | **1.00 (30)** |
| **storage (15)** | 1.00 (15) | 1.00 (15) | 1.00 (15) | 1.00 (15) | 1.00 (15) |
| **tpp (6)** | 1.00 (6) | 1.00 (6) | 1.00 (6) | 1.00 (6) | 1.00 (6) |
| **trucks-strips (10)** | **1.00 (10)** | 0.90 (9) | 0.90 (9) | 0.90 (9) | **1.00 (10)** |
| **elevators-opt08-strips (22)** | **1.00 (22)** | 0.82 (18) | 0.82 (18) | 0.82 (18) | 0.95 (21) |
| **openstacks-opt08-strips (20)** | **1.00 (20)** | 0.85 (17) | 0.85 (17) | 0.95 (19) | 0.95 (19) |
| **parcprinter-08-strips (22)** | 0.82 (18) | **0.95 (21)** | **0.95 (21)** | 0.82 (18) | 0.91 (20) |
| **pegsol-08-strips (28)** | **1.00 (28)** | 0.96 (27) | 0.96 (27) | 0.96 (27) | 0.96 (27) |
| **scanalyzer-08-strips (16)** | **0.94 (15)** | 0.81 (13) | 0.81 (13) | 0.81 (13) | **0.94 (15)** |
| **sokoban-opt08-strips (30)** | **1.00 (30)** | 0.83 (25) | 0.83 (25) | 0.83 (25) | 0.83 (25) |
| **transport-opt08-strips (12)** | 0.92 (11) | 0.92 (11) | 0.92 (11) | 0.92 (11) | 0.92 (11) |
| **woodworking-opt08-strips (19)** | 0.84 (16) | 0.74 (14) | 0.79 (15) | 0.74 (14) | **0.95 (18)** |
| **TOTAL** | **0.92 (639)** | 0.89 (614) | 0.89 (614) | 0.87 (602) | 0.91 (630) |

Table 14: Detailed per-domain normalized coverage using $h_{\text{LM-CUT}}$ and $h_{\text{LM-CUT}^+}$. Each line shows the normalized coverage in each domain, with the number of problems solved is shown in parentheses. Domains are grouped into domains with unit cost actions and high variance in coverage, domains with unit cost actions and low variance in coverage, and domains with non-uniform action costs, respectively.

| expansions | $h_{\text{LM-CUT}}$ | $h_{\text{LM-CUT}+}$ | $\max_h$ | $\text{rnd}_h$ | $\text{sel}_h$ |
|---|---|---|---|---|---|
| **airport (26)** | 1.16 | **1.0** | **1.0** | 1.04 | 1.16 |
| **freecell (12)** | 9.55 | **1.0** | **1.0** | 4.37 | 1.26 |
| **logistics00 (16)** | 1.0 | 1.0 | 1.0 | 1.0 | 1.0 |
| **mprime (21)** | 2.2 | 1.01 | **1.0** | 1.84 | **1.0** |
| **mystery (16)** | 1.69 | 1.01 | **1.0** | 1.52 | 1.32 |
| **pipesworld-tankage (8)** | 3.09 | 1.01 | **1.0** | 1.75 | 1.61 |
| **satellite (7)** | 3.66 | **0.98** | 1.0 | 2.39 | 1.51 |
| **zenotravel (12)** | 1.61 | 1.09 | **1.0** | 1.3 | 1.22 |
| **blocks (27)** | 1.02 | **1.0** | **1.0** | 1.01 | 1.02 |
| **depot (7)** | 7.53 | **1.0** | **1.0** | 4.07 | 1.25 |
| **driverlog (13)** | 1.71 | 1.02 | **1.0** | 1.36 | 1.49 |
| **grid (2)** | 4.03 | **1.0** | **1.0** | 1.9 | 1.28 |
| **gripper (6)** | 1.05 | **1.0** | **1.0** | 1.03 | 1.05 |
| **logistics98 (6)** | 1.08 | 1.05 | **1.0** | 1.06 | 1.06 |
| **miconic (140)** | 1.0 | 1.0 | 1.0 | 1.0 | 1.0 |
| **pathways (5)** | 1.22 | 1.02 | **1.0** | 1.13 | 1.22 |
| **pipesworld-notankage (16)** | 3.49 | 1.01 | **1.0** | 1.9 | 1.4 |
| **psr-small (48)** | 1.03 | **1.0** | **1.0** | 1.02 | 1.03 |
| **rovers (7)** | 1.66 | 1.01 | **1.0** | 1.28 | 1.3 |
| **schedule (27)** | 1.0 | 1.0 | 1.0 | 1.0 | 1.0 |
| **storage (15)** | 1.07 | **1.0** | **1.0** | 1.03 | 1.07 |
| **tpp (6)** | 1.56 | **1.0** | **1.0** | 1.16 | 1.56 |
| **trucks-strips (9)** | 1.32 | 1.04 | **1.0** | 1.14 | 1.26 |
| **elevators-opt08-strips (18)** | 1.75 | 1.09 | **1.0** | 1.4 | 1.72 |
| **openstacks-opt08-strips (17)** | 1.0 | 1.0 | 1.0 | 1.0 | 1.0 |
| **parcprinter-08-strips (17)** | 1.71 | **1.0** | **1.0** | 1.37 | **1.0** |
| **pegsol-08-strips (27)** | 1.33 | 1.01 | **1.0** | 1.15 | 1.2 |
| **scanalyzer-08-strips (13)** | 1.22 | 1.02 | **1.0** | 1.14 | 1.13 |
| **sokoban-opt08-strips (25)** | 1.04 | 1.04 | **1.0** | 1.01 | 1.03 |
| **transport-opt08-strips (11)** | 1.29 | 1.01 | **1.0** | 1.15 | 1.26 |
| **woodworking-opt08-strips (13)** | 1.45 | 1.06 | **1.0** | 1.26 | 1.12 |
| **GEOMETRIC MEAN** | 1.66 | 1.01 | **1.0** | 1.35 | 1.2 |

Table 15: Detailed per-domain expansions relative to $\max_h$ using $h_{\text{LM-CUT}}$ and $h_{\text{LM-CUT}+}$. Each row shows the geometric mean of the ratio of expanded nodes relative to $\max_h$. Domains are grouped into domains with unit cost actions and high variance in coverage, domains with unit cost actions and low variance in coverage, and domains with non-uniform action costs, respectively.

| coverage | $h_{LA}$ | $h_{\text{LM-CUT}}$ | $h_{\text{LM-CUT}^+}$ | $\max_h$ | $\text{rnd}_h$ | $\text{sel}_h$ |
|---|---|---|---|---|---|---|
| **airport (33)** | 0.91 (30) | 0.85 (28) | **0.94 (31)** | **0.94 (31)** | 0.79 (26) | 0.91 (30) |
| **freecell (58)** | **1.00 (58)** | 0.26 (15) | 0.22 (13) | 0.53 (31) | 0.26 (15) | 0.57 (33) |
| **logistics00 (21)** | **1.00 (21)** | 0.95 (20) | 0.81 (17) | 0.76 (16) | 0.95 (20) | 0.95 (20) |
| **mprime (24)** | 0.88 (21) | **1.00 (24)** | **1.00 (24)** | **1.00 (24)** | 0.75 (18) | 0.96 (23) |
| **mystery (17)** | 0.88 (15) | **1.00 (17)** | **1.00 (17)** | **1.00 (17)** | 0.71 (12) | **1.00 (17)** |
| **pipesworld-tankage (13)** | **1.00 (13)** | 0.92 (12) | 0.69 (9) | 0.69 (9) | 0.69 (9) | 0.85 (11) |
| **satellite (10)** | 0.70 (7) | 0.70 (7) | **0.90 (9)** | **0.90 (9)** | 0.70 (7) | 0.80 (8) |
| **zenotravel (13)** | 0.77 (10) | **1.00 (13)** | 0.92 (12) | 0.92 (12) | 0.69 (9) | 0.92 (12) |
| **blocks (28)** | 0.96 (27) | **1.00 (28)** | 0.96 (27) | 0.96 (27) | 0.96 (27) | **1.00 (28)** |
| **depot (7)** | 1.00 (7) | 1.00 (7) | 1.00 (7) | 1.00 (7) | 1.00 (7) | 1.00 (7) |
| **driverlog (14)** | **1.00 (14)** | 0.93 (13) | **1.00 (14)** | **1.00 (14)** | 0.93 (13) | 0.93 (13) |
| **grid (3)** | **1.00 (3)** | 0.67 (2) | 0.67 (2) | 0.67 (2) | 0.67 (2) | 0.67 (2) |
| **gripper (7)** | **1.00 (7)** | **1.00 (7)** | 0.86 (6) | 0.71 (5) | 0.86 (6) | **1.00 (7)** |
| **logistics98 (6)** | **1.00 (6)** | **1.00 (6)** | **1.00 (6)** | **1.00 (6)** | 0.83 (5) | **1.00 (6)** |
| **miconic (142)** | **1.00 (142)** | 0.99 (141) | 0.99 (140) | 0.99 (140) | 0.99 (140) | **1.00 (142)** |
| **pathways (5)** | 0.80 (4) | **1.00 (5)** | **1.00 (5)** | **1.00 (5)** | 0.80 (4) | **1.00 (5)** |
| **pipesworld-notankage (18)** | **1.00 (18)** | 0.94 (17) | 0.94 (17) | 0.94 (17) | 0.83 (15) | 0.94 (17) |
| **psr-small (49)** | **1.00 (49)** | **1.00 (49)** | 0.98 (48) | 0.98 (48) | 0.98 (48) | **1.00 (49)** |
| **rovers (8)** | **1.00 (8)** | 0.88 (7) | 0.88 (7) | 0.88 (7) | 0.88 (7) | **1.00 (8)** |
| **schedule (30)** | **1.00 (30)** | **1.00 (30)** | 0.90 (27) | 0.90 (27) | 0.93 (28) | **1.00 (30)** |
| **storage (15)** | 1.00 (15) | 1.00 (15) | 1.00 (15) | 1.00 (15) | 1.00 (15) | 1.00 (15) |
| **tpp (6)** | 1.00 (6) | 1.00 (6) | 1.00 (6) | 1.00 (6) | 1.00 (6) | 1.00 (6) |
| **trucks-strips (10)** | 0.90 (9) | **1.00 (10)** | 0.90 (9) | 0.90 (9) | 0.90 (9) | **1.00 (10)** |
| **elevators-opt08-strips (22)** | 0.77 (17) | **1.00 (22)** | 0.82 (18) | 0.82 (18) | 0.64 (14) | 0.95 (21) |
| **openstacks-opt08-strips (20)** | 0.90 (18) | **1.00 (20)** | 0.85 (17) | 0.80 (16) | 0.90 (18) | 0.80 (16) |
| **parcprinter-08-strips (22)** | 0.68 (15) | 0.82 (18) | **0.95 (21)** | **0.95 (21)** | 0.59 (13) | 0.86 (19) |
| **pegsol-08-strips (28)** | 0.96 (27) | **1.00 (28)** | 0.96 (27) | 0.96 (27) | 0.96 (27) | 0.96 (27) |
| **scanalyzer-08-strips (16)** | 0.56 (9) | **0.94 (15)** | 0.81 (13) | 0.81 (13) | 0.38 (6) | **0.94 (15)** |
| **sokoban-opt08-strips (30)** | 0.83 (25) | **1.00 (30)** | 0.83 (25) | 0.77 (23) | 0.90 (27) | 0.87 (26) |
| **transport-opt08-strips (12)** | **1.00 (12)** | 0.92 (11) | 0.92 (11) | 0.92 (11) | 0.92 (11) | 0.92 (11) |
| **woodworking-opt08-strips (19)** | 0.68 (13) | **0.84 (16)** | 0.74 (14) | 0.79 (15) | 0.74 (14) | 0.79 (15) |
| **TOTAL** | 0.91 (656) | **0.92 (639)** | 0.89 (614) | 0.89 (628) | 0.81 (578) | **0.92 (649)** |

Table 16: Detailed per-domain normalized coverage using $h_{LA}$, $h_{\text{LM-CUT}}$ and $h_{\text{LM-CUT}^+}$. Each line shows the normalized coverage in each domain, with the number of problems solved is shown in parentheses. Domains are grouped into domains with unit cost actions and high variance in coverage, domains with unit cost actions and low variance in coverage, and domains with non-uniform action costs, respectively.

| expansions | $h_{LA}$ | $h_{\text{LM-CUT}}$ | $h_{\text{LM-CUT}^+}$ | $\max_h$ | $\text{rnd}_h$ | $\text{sel}_h$ |
|---|---|---|---|---|---|---|
| **airport (26)** | 2.29 | 1.16 | **1.0** | **1.0** | 1.04 | 1.71 |
| **freecell (13)** | 1.22 | 417.8 | 47.65 | **1.0** | 45.83 | 6.73 |
| **logistics00 (16)** | 1.0 | 1.0 | 1.0 | **1.0** | 1.0 | 1.0 |
| **mprime (18)** | 9.21 | 2.74 | 1.21 | **1.0** | 4.26 | 1.99 |
| **mystery (13)** | 7.85 | 1.41 | 1.13 | **1.0** | 4.48 | 1.43 |
| **pipesworld-tankage (9)** | 2.68 | 5.08 | 1.38 | **1.0** | 2.27 | 1.93 |
| **satellite (7)** | 18.81 | 3.78 | 1.01 | **1.0** | 4.53 | 2.45 |
| **zenotravel (9)** | 7.26 | 1.23 | 1.1 | **1.0** | 3.07 | 2.45 |
| **blocks (27)** | 7.59 | 1.02 | **1.0** | **1.0** | 1.58 | 1.67 |
| **depot (7)** | 19.63 | 7.53 | 1.01 | **1.0** | 5.22 | 2.46 |
| **driverlog (13)** | 11.36 | 1.73 | 1.03 | **1.0** | 2.79 | 2.03 |
| **grid (2)** | 5.04 | 4.06 | 1.01 | **1.0** | 2.15 | 4.91 |
| **gripper (5)** | **1.0** | 1.06 | **1.0** | **1.0** | 1.02 | **1.0** |
| **logistics98 (5)** | 6.43 | 1.08 | 1.05 | **1.0** | 1.79 | 1.58 |
| **miconic (140)** | 1.0 | 1.0 | 1.0 | **1.0** | 1.0 | 1.0 |
| **pathways (4)** | 40.63 | 1.02 | **1.0** | **1.0** | 7.76 | **1.0** |
| **pipesworld-notankage (15)** | 3.09 | 4.29 | 1.13 | **1.0** | 2.35 | 2.53 |
| **psr-small (48)** | 1.31 | 1.03 | **1.0** | **1.0** | 1.1 | 1.24 |
| **rovers (7)** | 2.77 | 1.67 | 1.01 | **1.0** | 1.78 | 1.38 |
| **schedule (27)** | 1.09 | 1.0 | 1.0 | **1.0** | **0.99** | 1.09 |
| **storage (15)** | 2.3 | 1.07 | 1.01 | **1.0** | 1.33 | 1.58 |
| **tpp (6)** | 2.73 | 1.56 | **1.0** | **1.0** | 1.91 | 1.41 |
| **trucks-strips (9)** | 60.39 | 1.33 | 1.01 | **1.0** | 6.05 | 1.33 |
| **elevators-opt08-strips (14)** | 33.16 | 1.65 | 1.1 | **1.0** | 4.65 | 2.9 |
| **openstacks-opt08-strips (16)** | 1.17 | **1.0** | **1.0** | **1.0** | 1.03 | 1.07 |
| **parcprinter-08-strips (13)** | 45.31 | 2.02 | **1.0** | **1.0** | 5.91 | **1.0** |
| **pegsol-08-strips (27)** | 4.94 | 1.34 | 1.01 | **1.0** | 1.69 | 1.26 |
| **scanalyzer-08-strips (6)** | 24.13 | 1.5 | 1.05 | **1.0** | 5.5 | 1.87 |
| **sokoban-opt08-strips (21)** | 16.43 | 1.03 | 1.05 | **1.0** | 1.14 | 1.13 |
| **transport-opt08-strips (11)** | 15.5 | 1.29 | 1.01 | **1.0** | 2.66 | 1.82 |
| **woodworking-opt08-strips (11)** | 53.33 | 1.37 | **1.0** | **1.0** | 4.02 | 1.63 |
| **GEOMETRIC MEAN** | 6.1 | 1.91 | 1.18 | **1.0** | 2.56 | 1.67 |

Table 17: Detailed per-domain expansions relative to $\max_h$ using $h_{LA}$, $h_{\text{LM-CUT}}$ and $h_{\text{LM-CUT}^+}$. Each row shows the geometric mean of the ratio of expanded nodes relative to $\max_h$. Domains are grouped into domains with unit cost actions and high variance in coverage, domains with unit cost actions and low variance in coverage, and domains with non-uniform action costs, respectively.

| overhead | $h_{LA}/h_{\textbf{LM-CUT}}$ | $h_{LA}/h_{\textbf{LM-CUT}+}$ | $h_{\textbf{LM-CUT}}/h_{\textbf{LM-CUT}+}$ | All Three |
|---|---|---|---|---|
| **airport (28)** | 4% | 7% | 1% | 9% |
| **freecell (13)** | 4% | 8% | 13% | 1% |
| **logistics00 (20)** | 8% | 7% | 2% | 6% |
| **mprime (23)** | 7% | 7% | 6% | 3% |
| **mystery (17)** | 3% | 3% | 8% | 2% |
| **pipesworld-tankage (9)** | 11% | 11% | 10% | 5% |
| **satellite (7)** | 14% | 18% | 10% | 8% |
| **zenotravel (12)** | 15% | 35% | 26% | 21% |
| **blocks (26)** | 21% | 35% | 2% | 5% |
| **depot (7)** | 45% | 29% | 14% | 10% |
| **driverlog (13)** | 29% | 45% | 26% | 21% |
| **grid (2)** | 26% | 17% | 1% | 6% |
| **gripper (7)** | 13% | 13% | 5% | 22% |
| **logistics98 (6)** | 15% | 31% | 6% | 5% |
| **miconic (141)** | 1% | 4% | 3% | 4% |
| **pathways (5)** | 5% | 1% | 4% | 7% |
| **pipesworld-notankage (17)** | 22% | 17% | 20% | 22% |
| **psr-small (49)** | 8% | 11% | 3% | 12% |
| **rovers (7)** | 15% | 24% | 26% | 19% |
| **schedule (30)** | 13% | 13% | 5% | 24% |
| **storage (15)** | 18% | 12% | 2% | 10% |
| **tpp (6)** | 2% | 1% | 2% | 3% |
| **trucks-strips (9)** | 3% | 2% | 12% | 7% |
| **elevators-opt08-strips (16)** | 32% | 75% | 8% | 9% |
| **openstacks-opt08-strips (16)** | 15% | 9% | 10% | 23% |
| **parcprinter-08-strips (18)** | 2% | 6% | 1% | 5% |
| **pegsol-08-strips (27)** | 9% | 2% | 28% | 15% |
| **scanalyzer-08-strips (13)** | 2% | 4% | 10% | 1% |
| **sokoban-opt08-strips (24)** | 5% | 2% | 14% | 7% |
| **transport-opt08-strips (11)** | 12% | 23% | 7% | 3% |
| **woodworking-opt08-strips (14)** | 5% | 5% | 2% | 4% |
| **AVERAGE** | 12% | 15% | 9% | 10% |

Table 18: Selective max overhead. Each row lists the average percentage of time spent on learning and classification, out of the total time taken by selective max, in each domain, for each set of heuristics. Domains are grouped into domains with unit cost actions and high variance in coverage, domains with unit cost actions and low variance in coverage, and domains with non-uniform action costs, respectively.

| coverage | $\text{sel}_h^{\alpha=0.1}$ | $\text{sel}_h^{\alpha=0.5}$ | $\text{sel}_h^{\alpha=1}$ | $\text{sel}_h^{\alpha=1.5}$ | $\text{sel}_h^{\alpha=2}$ | $\text{sel}_h^{\alpha=3}$ | $\text{sel}_h^{\alpha=4}$ | $\text{sel}_h^{\alpha=5}$ |
|---|---|---|---|---|---|---|---|---|
| **airport (50)** | 30 | 30 | 30 | 30 | 30 | 30 | 30 | 30 |
| **freecell (80)** | 49 | 49 | 49 | 49 | 49 | 49 | 49 | 49 |
| **logistics00 (28)** | 21 | 21 | 21 | 21 | 21 | 21 | 21 | 21 |
| **mprime (35)** | **24** | **24** | **24** | **24** | 22 | 23 | 21 | 21 |
| **mystery (30)** | **17** | **17** | **17** | **17** | **17** | 16 | 15 | 15 |
| **pipesworld-tankage (50)** | 12 | 12 | 12 | 12 | 12 | 12 | **13** | **13** |
| **satellite (36)** | **8** | **8** | **8** | **8** | 7 | 7 | 7 | 7 |
| **zenotravel (20)** | **13** | **13** | **13** | **13** | 12 | 11 | 10 | 10 |
| **SUM** | **174** | **174** | **174** | **174** | 170 | 169 | 166 | 166 |

Table 19: Number of problems solved by selective max in each domain with varying values of hyper-parameter $\alpha$

| coverage | $\text{sel}_h^{\rho=0.51}$ | $\text{sel}_h^{\rho=0.6}$ | $\text{sel}_h^{\rho=0.7}$ | $\text{sel}_h^{\rho=0.8}$ | $\text{sel}_h^{\rho=0.9}$ | $\text{sel}_h^{\rho=0.99}$ |
|---|---|---|---|---|---|---|
| **airport (50)** | 30 | 30 | 30 | 30 | 30 | 30 |
| **freecell (80)** | 48 | **49** | **49** | **49** | **49** | **49** |
| **logistics00 (28)** | 21 | 21 | 21 | 21 | 21 | 21 |
| **mprime (35)** | 24 | 24 | 24 | 24 | 24 | 24 |
| **mystery (30)** | 17 | 17 | 17 | 17 | 17 | 17 |
| **pipesworld-tankage (50)** | 12 | 12 | 12 | 12 | 12 | 12 |
| **satellite (36)** | 8 | 8 | 8 | 8 | 8 | 8 |
| **zenotravel (20)** | 13 | 13 | 13 | 13 | 13 | 13 |
| **SUM** | 173 | **174** | **174** | **174** | **174** | **174** |

Table 20: Number of problems solved by selective max in each domain with varying values of confidence threshold $\rho$

Table 18 lists the average overhead of selective max in each domain, for each combination of two or more heuristics.

Tables 19, 20, 21, 22 and 23 list the number of problems solved in each domain, under various values for $\alpha$, $\rho$, $N$, sampling method and classifier, respectively.

| coverage | $\text{sel}_h^{N=10}$ | $\text{sel}_h^{N=100}$ | $\text{sel}_h^{N=1000}$ |
|---|---|---|---|
| **airport (50)** | 30 | 30 | 30 |
| **freecell (80)** | 47 | **49** | 46 |
| **logistics00 (28)** | 21 | 21 | 21 |
| **mprime (35)** | 24 | 24 | 24 |
| **mystery (30)** | 17 | 17 | 17 |
| **pipesworld-tankage (50)** | 12 | 12 | 12 |
| **satellite (36)** | 8 | 8 | 8 |
| **zenotravel (20)** | 13 | 13 | 13 |
| **SUM** | 172 | **174** | 171 |

Table 21: Number of problems solved by selective max in each domain with varying values of initial Sample Size $N$

| coverage | $sel_h^{PDB}$ | $sel_h^{P}$ | $sel_h^{UP}$ |
|---|---|---|---|
| **airport (50)** | 30 | 30 | 30 |
| **freecell (80)** | 49 | 53 | **55** |
| **logistics00 (28)** | 21 | 21 | 21 |
| **mprime (35)** | 24 | 24 | 24 |
| **mystery (30)** | 17 | 17 | 17 |
| **pipesworld-tankage (50)** | 12 | 12 | 12 |
| **satellite (36)** | 8 | 8 | 8 |
| **zenotravel (20)** | 13 | 13 | 13 |
| **SUM** | 174 | 178 | **180** |

Table 22: Number of problems solved by selective max in each domain with different sampling methods. $PDB$ is the sampling method of Haslum et al. (2007), $P$ is the biased probes sampling method, and $UP$ is the unbiased probes sampling method.

| coverage | $sel_h^{NB}$ | $sel_h^{AODE}$ | $sel_h^{ITI}$ | $sel_h^{3NN}$ | $sel_h^{5NN}$ |
|---|---|---|---|---|---|
| **airport (50)** | **30** | 25 | **30** | **30** | 28 |
| **freecell (80)** | **49** | **49** | 34 | 35 | 46 |
| **logistics00 (28)** | **21** | 20 | 20 | 20 | 20 |
| **mprime (35)** | **24** | **24** | **24** | **24** | 23 |
| **mystery (30)** | 17 | 17 | 17 | 17 | 17 |
| **pipesworld-tankage (50)** | **12** | **12** | **12** | **12** | 10 |
| **satellite (36)** | **8** | **8** | 7 | 7 | 6 |
| **zenotravel (20)** | **13** | **13** | 12 | **13** | 11 |
| **SUM** | **174** | 168 | 156 | 158 | 161 |

Table 23: Number of problems solved by selective max in each domain with different classifiers

| coverage | $\mathrm{sel}_h$ | $\mathrm{port}_{int}$ | $\mathrm{port}_{ctr}$ |
|---|---|---|---|
| **airport (33)** | **0.91** (30) | **0.91** (30) | **0.91** (30) |
| **freecell (58)** | 0.84 (49) | 0.91 (53) | **0.93** (54) |
| **logistics00 (21)** | **1.00** (21) | 0.95 (20) | **1.00** (21) |
| **mprime (24)** | **1.00** (24) | 0.96 (23) | 0.96 (23) |
| **mystery (17)** | 1.00 (17) | 1.12 (19) | **1.24** (21) |
| **pipesworld-tankage (13)** | 0.92 (12) | 0.92 (12) | **1.00** (13) |
| **satellite (10)** | **0.80** (8) | 0.70 (7) | 0.70 (7) |
| **zenotravel (13)** | **1.00** (13) | 0.92 (12) | 0.92 (12) |
| **blocks (28)** | **1.00** (28) | **1.00** (28) | **1.00** (28) |
| **depot (7)** | **1.00** (7) | **1.00** (7) | **1.00** (7) |
| **driverlog (14)** | **1.00** (14) | **1.00** (14) | **1.00** (14) |
| **grid (3)** | **0.67** (2) | **0.67** (2) | **0.67** (2) |
| **gripper (7)** | **1.00** (7) | **1.00** (7) | **1.00** (7) |
| **logistics98 (6)** | **1.00** (6) | **1.00** (6) | **1.00** (6) |
| **miconic (142)** | **1.00** (142) | **1.00** (142) | **1.00** (142) |
| **pathways (5)** | **1.00** (5) | **1.00** (5) | **1.00** (5) |
| **pipesworld-notankage (18)** | 0.94 (17) | **1.00** (18) | **1.00** (18) |
| **psr-small (49)** | **1.00** (49) | **1.00** (49) | **1.00** (49) |
| **rovers (8)** | **1.00** (8) | **1.00** (8) | **1.00** (8) |
| **schedule (30)** | **1.00** (30) | **1.00** (30) | **1.00** (30) |
| **storage (15)** | **1.00** (15) | **1.00** (15) | **1.00** (15) |
| **tpp (6)** | **1.00** (6) | **1.00** (6) | **1.00** (6) |
| **trucks-strips (10)** | **1.00** (10) | 0.90 (9) | **1.00** (10) |
| **elevators-opt08-strips (22)** | **1.00** (22) | 0.82 (18) | 0.86 (19) |
| **openstacks-opt08-strips (20)** | 0.90 (18) | 0.90 (18) | **0.95** (19) |
| **parcprinter-08-strips (22)** | **0.82** (18) | **0.82** (18) | **0.82** (18) |
| **pegsol-08-strips (28)** | **0.96** (27) | **0.96** (27) | **0.96** (27) |
| **scanalyzer-08-strips (16)** | 0.94 (15) | 0.81 (13) | **1.00** (16) |
| **sokoban-opt08-strips (30)** | **0.97** (29) | **0.97** (29) | **0.97** (29) |
| **transport-opt08-strips (12)** | 0.92 (11) | 0.92 (11) | **1.00** (12) |
| **woodworking-opt08-strips (19)** | **0.89** (17) | 0.84 (16) | **0.89** (17) |
| **TOTAL** | 0.95 (677) | 0.94 (672) | **0.96 (685)** |

Table 24: Detailed coverage of portfolio using $h_{LA}$ / $h_{\text{LM-CUT}}$. Number of problems solved by selective max ($\mathrm{sel}_h$), a simulated interruptible portfolio ($\mathrm{port}_{int}$), and a simulated contract anytime portfolio ($\mathrm{port}_{ctr}$) in each domain using heuristics $h_{LA}$ / $h_{\text{LM-CUT}}$. Domains are grouped into domains with unit cost actions and high variance in coverage, domains with unit cost actions and low variance in coverage, and domains with non-uniform action costs, respectively.

| coverage | $\text{sel}_h$ | $\text{port}_{int}$ | $\text{port}_{ctr}$ |
|---|---|---|---|
| **airport (33)** | **0.91 (30)** | **0.91 (30)** | **0.91 (30)** |
| **freecell (58)** | 0.71 (41) | 0.91 (53) | **0.93 (54)** |
| **logistics00 (21)** | **1.00 (21)** | 0.95 (20) | **1.00 (21)** |
| **mprime (24)** | **1.00 (24)** | **1.00 (24)** | **1.00 (24)** |
| **mystery (17)** | 1.00 (17) | 1.12 (19) | **1.18 (20)** |
| **pipesworld-tankage (13)** | 0.69 (9) | 0.92 (12) | **1.00 (13)** |
| **satellite (10)** | **1.00 (10)** | 0.80 (8) | 0.80 (8) |
| **zenotravel (13)** | **0.92 (12)** | 0.85 (11) | 0.85 (11) |
| **blocks (28)** | 0.93 (26) | 0.93 (26) | **0.96 (27)** |
| **depot (7)** | **1.00 (7)** | **1.00 (7)** | **1.00 (7)** |
| **driverlog (14)** | 0.93 (13) | **1.00 (14)** | **1.00 (14)** |
| **grid (3)** | **0.67 (2)** | **0.67 (2)** | **0.67 (2)** |
| **gripper (7)** | **1.00 (7)** | **1.00 (7)** | **1.00 (7)** |
| **logistics98 (6)** | **1.00 (6)** | **1.00 (6)** | **1.00 (6)** |
| **miconic (142)** | **1.00 (142)** | **1.00 (142)** | **1.00 (142)** |
| **pathways (5)** | **1.00 (5)** | **1.00 (5)** | **1.00 (5)** |
| **pipesworld-notankage (18)** | 0.94 (17) | **1.00 (18)** | **1.00 (18)** |
| **psr-small (49)** | **1.00 (49)** | **1.00 (49)** | **1.00 (49)** |
| **rovers (8)** | **1.00 (8)** | **1.00 (8)** | **1.00 (8)** |
| **schedule (30)** | **1.00 (30)** | **1.00 (30)** | **1.00 (30)** |
| **storage (15)** | **1.00 (15)** | **1.00 (15)** | **1.00 (15)** |
| **tpp (6)** | **1.00 (6)** | **1.00 (6)** | **1.00 (6)** |
| **trucks-strips (10)** | **0.90 (9)** | 0.70 (7) | 0.80 (8) |
| **elevators-opt08-strips (22)** | 0.73 (16) | **0.82 (18)** | **0.82 (18)** |
| **openstacks-opt08-strips (20)** | **0.85 (17)** | **0.85 (17)** | **0.85 (17)** |
| **parcprinter-08-strips (22)** | **1.00 (22)** | **1.00 (22)** | **1.00 (22)** |
| **pegsol-08-strips (28)** | **0.96 (27)** | **0.96 (27)** | **0.96 (27)** |
| **scanalyzer-08-strips (16)** | 0.81 (13) | 0.75 (12) | **0.94 (15)** |
| **sokoban-opt08-strips (30)** | 0.80 (24) | **0.83 (25)** | **0.83 (25)** |
| **transport-opt08-strips (12)** | 0.92 (11) | 0.92 (11) | **1.00 (12)** |
| **woodworking-opt08-strips (19)** | **0.79 (15)** | **0.79 (15)** | **0.79 (15)** |
| **TOTAL** | 0.92 (651) | 0.93 (666) | **0.94 (676)** |

Table 25: Detailed coverage of portfolio using $h_{LA}$ / $h_{\text{LM-CUT}^+}$. Number of problems solved by selective max ($\text{sel}_h$), a simulated interruptible portfolio ($\text{port}_{int}$), and a simulated contract anytime portfolio ($\text{port}_{ctr}$) in each domain using heuristics $h_{LA}$ / $h_{\text{LM-CUT}^+}$. Domains are grouped into domains with unit cost actions and high variance in coverage, domains with unit cost actions and low variance in coverage, and domains with non-uniform action costs, respectively.

| coverage | $\text{sel}_h$ | $\text{port}_{int}$ | $\text{port}_{ctr}$ |
|---|---|---|---|
| **airport (33)** | 0.85 (28) | **0.88 (29)** | **0.88 (29)** |
| **freecell (58)** | 0.22 (13) | 0.24 (14) | **0.26 (15)** |
| **logistics00 (21)** | **0.95 (20)** | **0.95 (20)** | **0.95 (20)** |
| **mprime (24)** | **1.00 (24)** | **1.00 (24)** | **1.00 (24)** |
| **mystery (17)** | 0.94 (16) | 1.18 (20) | **1.24 (21)** |
| **pipesworld-tankage (13)** | 0.69 (9) | 0.69 (9) | **0.85** (11) |
| **satellite (10)** | **0.80 (8)** | **0.80 (8)** | **0.80 (8)** |
| **zenotravel (13)** | **0.92 (12)** | **0.92 (12)** | **0.92 (12)** |
| **blocks (28)** | **1.00 (28)** | **1.00 (28)** | **1.00 (28)** |
| **depot (7)** | **1.00 (7)** | **1.00 (7)** | **1.00 (7)** |
| **driverlog (14)** | **1.00 (14)** | 0.93 (13) | 0.93 (13) |
| **grid (3)** | **0.67 (2)** | **0.67 (2)** | **0.67 (2)** |
| **gripper (7)** | **1.00 (7)** | 0.86 (6) | **1.00 (7)** |
| **logistics98 (6)** | **1.00 (6)** | **1.00 (6)** | **1.00 (6)** |
| **miconic (142)** | **0.99 (141)** | 0.99 (140) | **0.99 (141)** |
| **pathways (5)** | **1.00 (5)** | **1.00 (5)** | **1.00 (5)** |
| **pipesworld-notankage (18)** | **0.94 (17)** | 0.89 (16) | **0.94 (17)** |
| **psr-small (49)** | **1.00 (49)** | **1.00 (49)** | **1.00 (49)** |
| **rovers (8)** | **0.88 (7)** | **0.88 (7)** | **0.88 (7)** |
| **schedule (30)** | **1.00 (30)** | 0.93 (28) | **1.00 (30)** |
| **storage (15)** | **1.00 (15)** | **1.00 (15)** | **1.00 (15)** |
| **tpp (6)** | **1.00 (6)** | **1.00 (6)** | **1.00 (6)** |
| **trucks-strips (10)** | **1.00 (10)** | 0.90 (9) | **1.00** (10) |
| **elevators-opt08-strips (22)** | **0.95 (21)** | 0.82 (18) | 0.86 (19) |
| **openstacks-opt08-strips (20)** | **0.95 (19)** | 0.90 (18) | **0.95 (19)** |
| **parcprinter-08-strips (22)** | 0.91 (20) | **1.00 (22)** | **1.00 (22)** |
| **pegsol-08-strips (28)** | **0.96 (27)** | **0.96 (27)** | **0.96 (27)** |
| **scanalyzer-08-strips (16)** | **0.94 (15)** | 0.81 (13) | **0.94 (15)** |
| **sokoban-opt08-strips (30)** | 0.83 (25) | **0.93 (28)** | **0.93 (28)** |
| **transport-opt08-strips (12)** | **0.92 (11)** | **0.92 (11)** | **0.92 (11)** |
| **woodworking-opt08-strips (19)** | **0.95 (18)** | 0.79 (15) | 0.84 (16) |
| **TOTAL** | 0.91 (630) | 0.90 (625) | **0.93 (640)** |

Table 26: Detailed coverage of portfolio using $h_{\text{LM-CUT}}$ / $h_{\text{LM-CUT}^+}$. Number of problems solved by selective max ($\text{sel}_h$), a simulated interruptible portfolio ($\text{port}_{int}$), and a simulated contract anytime portfolio ($\text{port}_{ctr}$) in each domain using heuristics $h_{\text{LM-CUT}}$ / $h_{\text{LM-CUT}^+}$. Domains are grouped into domains with unit cost actions and high variance in coverage, domains with unit cost actions and low variance in coverage, and domains with non-uniform action costs, respectively.

| coverage | $\mathrm{sel}_h$ | $\mathrm{port}_{int}$ | $\mathrm{port}_{ctr}$ |
|---|---|---|---|
| **airport (33)** | **0.91 (30)** | **0.91 (30)** | **0.91 (30)** |
| **freecell (58)** | 0.57 (33) | 0.91 (53) | **0.93 (54)** |
| **logistics00 (21)** | 0.95 (20) | 0.95 (20) | **1.00 (21)** |
| **mprime (24)** | 0.96 (23) | **1.00 (24)** | **1.00 (24)** |
| **mystery (17)** | 1.00 (17) | **1.18 (20)** | **1.18 (20)** |
| **pipesworld-tankage (13)** | 0.85 (11) | **0.92 (12)** | **0.92 (12)** |
| **satellite (10)** | **0.80 (8)** | **0.80 (8)** | **0.80 (8)** |
| **zenotravel (13)** | **0.92 (12)** | **0.92 (12)** | **0.92 (12)** |
| **blocks (28)** | **1.00 (28)** | **1.00 (28)** | **1.00 (28)** |
| **depot (7)** | **1.00 (7)** | **1.00 (7)** | **1.00 (7)** |
| **driverlog (14)** | 0.93 (13) | **1.00 (14)** | **1.00 (14)** |
| **grid (3)** | **0.67 (2)** | **0.67 (2)** | **0.67 (2)** |
| **gripper (7)** | **1.00 (7)** | **1.00 (7)** | **1.00 (7)** |
| **logistics98 (6)** | **1.00 (6)** | **1.00 (6)** | **1.00 (6)** |
| **miconic (142)** | **1.00 (142)** | **1.00 (142)** | **1.00 (142)** |
| **pathways (5)** | **1.00 (5)** | **1.00 (5)** | **1.00 (5)** |
| **pipesworld-notankage (18)** | 0.94 (17) | **1.00 (18)** | **1.00 (18)** |
| **psr-small (49)** | **1.00 (49)** | **1.00 (49)** | **1.00 (49)** |
| **rovers (8)** | **1.00 (8)** | **1.00 (8)** | **1.00 (8)** |
| **schedule (30)** | **1.00 (30)** | **1.00 (30)** | **1.00 (30)** |
| **storage (15)** | **1.00 (15)** | **1.00 (15)** | **1.00 (15)** |
| **tpp (6)** | **1.00 (6)** | **1.00 (6)** | **1.00 (6)** |
| **trucks-strips (10)** | **1.00 (10)** | 0.90 (9) | 0.90 (9) |
| **elevators-opt08-strips (22)** | **0.95 (21)** | 0.82 (18) | 0.86 (19) |
| **openstacks-opt08-strips (20)** | 0.80 (16) | 0.90 (18) | **0.95 (19)** |
| **parcprinter-08-strips (22)** | 0.86 (19) | **1.00 (22)** | **1.00 (22)** |
| **pegsol-08-strips (28)** | **0.96 (27)** | **0.96 (27)** | **0.96 (27)** |
| **scanalyzer-08-strips (16)** | **0.94 (15)** | 0.81 (13) | 0.81 (13) |
| **sokoban-opt08-strips (30)** | 0.87 (26) | **0.97 (29)** | **0.97 (29)** |
| **transport-opt08-strips (12)** | **0.92 (11)** | **0.92 (11)** | **0.92 (11)** |
| **woodworking-opt08-strips (19)** | 0.79 (15) | 0.84 (16) | **0.89 (17)** |
| **TOTAL** | 0.92 (649) | **0.95 (679)** | **0.95 (684)** |

Table 27: Detailed coverage of portfolio using $h_{LA}$ / $h_{\mathrm{LM\text{-}CUT}}$ / $h_{\mathrm{LM\text{-}CUT}^+}$. Number of problems solved by selective max ($\mathrm{sel}_h$), a simulated interruptible portfolio ($\mathrm{port}_{int}$), and a simulated contract anytime portfolio ($\mathrm{port}_{ctr}$) in each domain using heuristics $h_{LA}$ / $h_{\mathrm{LM\text{-}CUT}}$ / $h_{\mathrm{LM\text{-}CUT}^+}$. Domains are grouped into domains with unit cost actions and high variance in coverage, domains with unit cost actions and low variance in coverage, and domains with non-uniform action costs, respectively.

Tables 24, 25, 26 and 27 list the normalized coverage in each domain for selective max, and for the simulated contract and interruptible sequential portfolios.

## References

Arfaee, S. J., Zilles, S., & Holte, R. C. (2010). Bootstrap learning of heuristic functions. In Felner, A., & Sturtevant, N. (Eds.), *Proceedings of the Third Annual Symposium on Combinatorial Search (SoCS 2010)*, pp. 52–60. AAAI Press.

Bäckström, C., & Klein, I. (1991). Planning in polynomial time: the SAS-PUBS class. *Computational Intelligence*, *7*(3), 181–197.

Bäckström, C., & Nebel, B. (1995). Complexity results for SAS$^+$ planning. *Computational Intelligence*, *11*(4), 625–655.

Bayardo Jr., R. J., & Schrag, R. (1997). Using CSP look-back techniques to solve real-world SAT instances. In Kuipers, B., & Webber, B. L. (Eds.), *Proceedings of the Fourteenth National Conference on Artificial Intelligence (AAAI 1997)*, pp. 203–208. AAAI Press.

Bonet, B., & Helmert, M. (2010). Strengthening landmark heuristics via hitting sets. In Coelho, H., Studer, R., & Wooldridge, M. (Eds.), *Proceedings of the 19th European Conference on Artificial Intelligence (ECAI 2010)*, pp. 329–334. IOS Press.

Bonet, B., Loerincs, G., & Geffner, H. (1997). A robust and fast action selection mechanism for planning. In Kuipers, B., & Webber, B. L. (Eds.), *Proceedings of the Fourteenth National Conference on Artificial Intelligence (AAAI 1997)*, pp. 714–719. AAAI Press.

Botea, A., Enzenberger, M., Müller, M., & Schaeffer, J. (2005). Macro-FF: Improving AI planning with automatically learned macro-operators. *Journal of Artificial Intelligence Research*, *24*, 581–621.

Brafman, R., & Shani, G. (2012). A multi-path compilation approach to contingent planning. In Hoffmann, J., & Selman, B. (Eds.), *Proceedings of the Twenty-Sixth AAAI Conference on Artificial Intelligence (AAAI 2012)*, pp. 9–15. AAAI Press.

Burke, E., Kendall, G., Newall, J., Hart, E., Ross, P., & Schulenburg, S. (2003). Hyper-Heuristics: An Emerging Direction in Modern Search Technology. In *Handbook of Metaheuristics*, International Series in Operations Research & Management Science, chap. 16, pp. 457–474.

Coles, A., & Smith, A. (2007). Marvin: A heuristic search planner with online macro-action learning. *Journal of Artificial Intelligence Research*, *28*, 119–156.

Cover, T. M., & Hart, P. E. (1967). Nearest neighbor pattern classification. *IEEE Transactions on Information Theory*, *13*(1), 21 – 27.

de la Rosa, T., Jiménez, S., & Borrajo, D. (2008). Learning relational decision trees for guiding heuristic planning. In Rintanen, J., Nebel, B., Beck, J. C., & Hansen, E. (Eds.), *Proceedings of the Eighteenth International Conference on Automated Planning and Scheduling (ICAPS 2008)*, pp. 60–67. AAAI Press.

Domshlak, C., Karpas, E., & Markovitch, S. (2010). To max or not to max: Online learning for speeding up optimal planning. In Fox, M., & Poole, D. (Eds.), *Proceedings of the Twenty-Fourth AAAI Conference on Artificial Intelligence (AAAI 2010)*, pp. 1071–1076. AAAI Press.

Fern, A. (2010). Speedup learning. In Sammut, C., & Webb, G. I. (Eds.), *Encyclopedia of Machine Learning*, pp. 907–911. Springer.

Fern, A., Khardon, R., & Tadepalli, P. (2011). The first learning track of the international planning competition. *Machine Learning*, *84*(1-2), 81–107.

Fikes, R. E., Hart, P. E., & Nilsson, N. J. (1972). Learning and executing generalized robot plans. *Artificial Intelligence*, *3*, 251–288.

Fikes, R. E., & Nilsson, N. J. (1971). STRIPS: A new approach to the application of theorem proving to problem solving. *Artificial Intelligence*, *2*, 189–208.

Finkelstein, L., & Markovitch, S. (1998). A selective macro-learning algorithm and its application to the NxN sliding-tile puzzle. *Journal of Artificial Intelligence Research*, *8*, 223–263.

García-Olaya, A., Jiménez, S., & Linares López, C. (2011). The 2011 international planning competition. Tech. rep., Universidad Carlos III de Madrid. http://hdl.handle.net/10016/11710.

Geffner, H. (2010). The model-based approach to autonomous behavior: A personal view. In Fox, M., & Poole, D. (Eds.), *Proceedings of the Twenty-Fourth AAAI Conference on Artificial Intelligence (AAAI 2010)*, pp. 1709–1712. AAAI Press.

Haslum, P., Botea, A., Helmert, M., Bonet, B., & Koenig, S. (2007). Domain-independent construction of pattern database heuristics for cost-optimal planning. In Holte, R. C., & Howe, A. E. (Eds.), *Proceedings of the Twenty-Second AAAI Conference on Artificial Intelligence (AAAI 2007)*, pp. 1007–1012. AAAI Press.

Helmert, M. (2006). The Fast Downward planning system. *Journal of Artificial Intelligence Research*, *26*, 191–246.

Helmert, M., & Domshlak, C. (2009). Landmarks, critical paths and abstractions: What's the difference anyway?. In Gerevini, A., Howe, A., Cesta, A., & Refanidis, I. (Eds.), *Proceedings of the Nineteenth International Conference on Automated Planning and Scheduling (ICAPS 2009)*, pp. 162–169. AAAI Press.

Helmert, M., Haslum, P., & Hoffmann, J. (2007). Flexible abstraction heuristics for optimal sequential planning. In Boddy, M., Fox, M., & Thiébaux, S. (Eds.), *Proceedings of the Seventeenth International Conference on Automated Planning and Scheduling (ICAPS 2007)*, pp. 176–183. AAAI Press.

Helmert, M., & Röger, G. (2008). How good is almost perfect?. In Fox, D., & Gomes, C. P. (Eds.), *Proceedings of the Twenty-Third AAAI Conference on Artificial Intelligence (AAAI 2008)*, pp. 944–949. AAAI Press.

Helmert, M., Röger, G., & Karpas, E. (2011). Fast Downward Stone Soup: A baseline for building planner portfolios. In *ICAPS 2011 Workshop on Planning and Learning*, pp. 28–35.

Hoffmann, J., & Nebel, B. (2001). The FF planning system: Fast plan generation through heuristic search. *Journal of Artificial Intelligence Research*, *14*, 253–302.

Hutter, F., Hoos, H. H., Leyton-Brown, K., & Stützle, T. (2009). ParamILS: An automatic algorithm configuration framework. *Journal of Artificial Intelligence Research*, *36*, 267–306.

Karpas, E., & Domshlak, C. (2009). Cost-optimal planning with landmarks. In Boutilier, C. (Ed.), *Proceedings of the 21st International Joint Conference on Artificial Intelligence (IJCAI 2009)*, pp. 1728–1733.

Katz, M., & Domshlak, C. (2010). Implicit abstraction heuristics. *Journal of Artificial Intelligence Research*, *39*, 51–126.

Kautz, H., & Selman, B. (1992). Planning as satisfiability. In Neumann, B. (Ed.), *Proceedings of the 10th European Conference on Artificial Intelligence (ECAI 1992)*, pp. 359–363. John Wiley and Sons.

Keyder, E., & Geffner, H. (2009). Soft goals can be compiled away. *Journal of Artificial Intelligence Research*, *36*, 547–556.

Marques-Silva, J. P., & Sakallah, K. A. (1996). GRASP - a new search algorithm for satisfiability. In *Proceedings of the 1996 IEEE/ACM International Conference on Computer-Aided Design (ICCAD 1996)*, pp. 220–227.

Minton, S. (1994). *Machine Learning Methods for Planning*. Morgan Kaufmann Publishers Inc.

Nissim, R., Hoffmann, J., & Helmert, M. (2011). Computing perfect heuristics in polynomial time: On bisimulation and merge-and-shrink abstraction in optimal planning. In Walsh, T. (Ed.), *Proceedings of the 22nd International Joint Conference on Artificial Intelligence (IJCAI'11)*, pp. 1983–1990. AAAI Press/IJCAI.

Palacios, H., & Geffner, H. (2009). Compiling uncertainty away in conformant planning problems with bounded width. *Journal of Artificial Intelligence Research*, *35*, 623–675.

Pearl, J. (1984). *Heuristics: Intelligent Search Strategies for Computer Problem Solving*. Addison-Wesley.

Pednault, E. P. D. (1989). ADL: Exploring the middle ground between STRIPS and the situation calculus. In Brachman, R. J., Levesque, H. J., & Reiter, R. (Eds.), *Proceedings of the First International Conference on Principles of Knowledge Representation and Reasoning (KR 1989)*, pp. 324–332. Morgan Kaufmann.

Rendell, L. A. (1983). A new basis for state-space learning systems and a successful implementation. *Artificial Intelligence*, *20*(4), 369–392.

Rintanen, J., Heljanko, K., & Niemelä, I. (2006). Planning as satisfiability: Parallel plans and algorithms for plan search. *Artificial Intelligence*, *170*(12–13), 1031–1080.

Russell, S. J., & Zilberstein, S. (1991). Composing real-time systems. In Mylopoulos, J., & Reiter, R. (Eds.), *Proceedings of the 12th International Joint Conference on Artificial Intelligence (IJCAI 1991)*, pp. 212–217. Morgan Kaufmann.

Schiex, T., & Verfaillie, G. (1993). Nogood recording for static and dynamic constraint satisfaction problems. *Journal of Artificial Intelligence Research*, *3*, 48–55.

Thayer, J. T., Dionne, A. J., & Ruml, W. (2011). Learning inadmissible heuristics during search. In Bacchus, F., Domshlak, C., Edelkamp, S., & Helmert, M. (Eds.), *Proceedings of the Twenty-First International Conference on Automated Planning and Scheduling (ICAPS 2011)*, pp. 250–257. AAAI Press.

Utgoff, P. E., Berkman, N. C., & Clouse, J. A. (1997). Decision tree induction based on efficient tree restructuring. *Machine Learning*, *29*(1), 5–44.

Webb, G. I., Boughton, J. R., & Wang, Z. (2005). Not so naive Bayes: Aggregating one-dependence estimators. *Machine Learning*, *58*(1), 5–24.

Yoon, S., Fern, A., & Givan, R. (2007). FF-Replan: A baseline for probabilistic planning. In Boddy, M., Fox, M., & Thiébaux, S. (Eds.), *Proceedings of the Seventeenth International Conference on Automated Planning and Scheduling (ICAPS 2007)*, pp. 352–359. AAAI Press.

Yoon, S., Fern, A., & Givan, R. (2008). Learning control knowledge for forward search planning. *Journal of Machine Learning Research*, *9*, 683–718.

Zimmerman, T., & Kambhampati, S. (2003). Learning-assisted automated planning: looking back, taking stock, going forward. *AI Magazine*, *24*, 73–96.